%% file: main.tex
\documentclass[sigconf,screen]{acmart}
\AtBeginDocument{%
  }

\copyrightyear{2026}
\acmYear{2026}
\setcopyright{cc}
\setcctype{by}
\acmConference[MM '26]{Proceedings of the 34th ACM International Conference on Multimedia}{November 10--14, 2026}{Rio de Janeiro, Brazil}
\acmBooktitle{Proceedings of the 34th ACM International Conference on Multimedia (MM '26), November 10--14, 2026, Rio de Janeiro, Brazil}
\acmDOI{10.1145/3767308.3836027}
\acmISBN{979-8-4007-2213-4/2026/11}
\acmSubmissionID{mfp6189}

\usepackage{float}
\usepackage{subfig}
\usepackage{makecell}
\usepackage{multicol}
\usepackage{multirow}
\usepackage{enumitem}
\usepackage{xspace}
\usepackage{stfloats}
\usepackage{mathtools}

\AtEndPreamble{
    \usepackage[capitalize]{cleveref}
    \crefname{section}{Sec.}{Secs.}
    \Crefname{section}{Section}{Sections}
    \Crefname{table}{Table}{Tables}
    \crefname{table}{Tab.}{Tabs.}
}

\newcommand{\mask}{\mathtt{[mask]}}
\newcommand{\method}{MIGM-Shortcut\xspace}
\newcommand{\dimo}{Di$\mathtt{[M]}$O\xspace}

\begin{document}

\title{Accelerating Masked Image Generation by Learning Controlled Latent Dynamics}

\input{secs/authors}

\renewcommand{\shortauthors}{Kaiwen Zhu et al.}

\input{secs/0-abs-v2}


%   \includegraphics[width=\textwidth]{samples/sampleteaser}


\maketitle

\input{secs/1-intro-v2}
\input{secs/2-related_work}

\input{secs/3-method}

\input{secs/4-exp}

\input{secs/5-conclusion}

\begin{acks}
This work is supported by Shanghai AI Laboratory.
\end{acks}

\bibliographystyle{ACM-Reference-Format}
\balance
\bibliography{main}

\clearpage
\appendix
\input{supp/feature_layer}
\input{supp/attempts}
\input{supp/more_eval}
\input{supp/more_visualization}
\input{supp/fail}
\input{supp/ablation_tab}

\input{supp/theoretical_support}
\input{supp/more_related_work}

\end{document}

%% file: secs/authors.tex
\author{Kaiwen Zhu}
% \orcid{0009-0002-5195-5707}
\authornote{This work was done during his internship at Shanghai AI Laboratory.}
% \authornote{Work done during internship at Shanghai AI Laboratory. \textsuperscript{\textdagger} Corresponding author.}
\affiliation{%
  \institution{Shanghai Jiao Tong University}
  \city{Shanghai}
  \country{China}
}
\affiliation{%
  \institution{Shanghai AI Laboratory}
  \city{Shanghai}
  \country{China}
}
\email{sqzhukaiwen@sjtu.edu.cn}

\author{Quansheng Zeng}
% \orcid{0009-0006-7026-162X}
\affiliation{%
  \institution{Shanghai AI Laboratory}
  \city{Shanghai}
  \country{China}
}
\affiliation{%
  \institution{Shanghai Innovation Institute}
  \city{Shanghai}
  \country{China}
}
\affiliation{%
  \institution{Nankai University}
  \city{Tianjin}
  \country{China}
}

\author{Yuandong Pu}
% \orcid{0000-0003-3764-0370}
\affiliation{%
  \institution{Shanghai Jiao Tong University}
  \city{Shanghai}
  \country{China}
}
\affiliation{%
  \institution{Shanghai AI Laboratory}
  \city{Shanghai}
  \country{China}
}

\author{Shuo Cao}
% \orcid{0009-0002-1507-8372}
\affiliation{%
  \institution{Shanghai AI Laboratory}
  \city{Shanghai}
  \country{China}
}
\affiliation{%
  \institution{University of Science and Technology of China}
  \city{Hefei}
  \country{China}
}

\author{Xiaohui Li}
% \orcid{0009-0005-9306-3514}
\affiliation{%
  \institution{Shanghai Jiao Tong University}
  \city{Shanghai}
  \country{China}
}
\affiliation{%
  \institution{Shanghai AI Laboratory}
  \city{Shanghai}
  \country{China}
}

\author{Yi Xin}
% \orcid{0009-0004-1113-6963}
\affiliation{%
  \institution{Shanghai AI Laboratory}
  \city{Shanghai}
  \country{China}
}
\affiliation{%
  \institution{Shanghai Innovation Institute}
  \city{Shanghai}
  \country{China}
}
\affiliation{%
  \institution{Nanjing University}
  \city{Nanjing}
  \country{China}
}

\author{Qi Qin}
% \orcid{0000-0003-2294-4005}
\affiliation{%
  \institution{Shanghai AI Laboratory}
  \city{Shanghai}
  \country{China}
}
\affiliation{%
  \institution{The University of Sydney}
  \city{Sydney}
  \country{Australia}
}

\author{Jiayang Li}
% \orcid{0009-0009-4267-6897}
\affiliation{%
  \institution{Shanghai AI Laboratory}
  \city{Shanghai}
  \country{China}
}
\affiliation{%
  \institution{Peking University}
  \city{Beijing}
  \country{China}
}

\author{Juncheng Yan}
% \orcid{0009-0001-4092-2435}
\affiliation{%
  \institution{Tsinghua University}
  \city{Beijing}
  \country{China}
}
\affiliation{%
  \institution{Shanghai AI Laboratory}
  \city{Shanghai}
  \country{China}
}

\author{Yu Qiao}
% \orcid{0000-0002-1889-2567}
\affiliation{%
  \institution{Shanghai AI Laboratory}
  \city{Shanghai}
  \country{China}
}
\affiliation{%
  \institution{Shanghai Innovation Institute}
  \city{Shanghai}
  \country{China}
}

\author{Jinjin Gu}
% \orcid{0000-0002-4389-6236}
\affiliation{%
  \institution{INSAIT, Sofia University ``St. Kliment Ohridski''}
  \city{Sofia}
  \country{Bulgaria}
}

\author{Yihao Liu}
% \orcid{0000-0001-9874-0602}
% \author{Yihao Liu$^\dagger$}
% \authornote{Corresponding author.}
\correspondingauthor
% \authornotemark[2]
\affiliation{%
  \institution{Shanghai AI Laboratory}
  \city{Shanghai}
  \country{China}
}
\email{liuyihao@pjlab.org.cn}

%% file: secs/0-abs-v2.tex
\begin{abstract}
Masked Image Generation Models (MIGMs) have achieved great success, yet their efficiency is hampered by the multiple steps of bi-directional attention.
In fact, there exists notable redundancy in their computation: when sampling discrete tokens, the rich semantics contained in the continuous features are lost. 
Some existing works attempt to cache the features to approximate future features. However, they exhibit considerable approximation error under aggressive acceleration settings.
We attribute this to their limited expressivity and the failure to account for sampling information.
To fill this gap, we propose learning a lightweight model that incorporates both previous features and sampled tokens, and regresses the average velocity field of feature evolution. 
The model has moderate complexity that suffices to capture the subtle dynamics while keeping lightweight compared to the original base model.
We apply our method to two representative MIGMs and tasks. In particular, on the state-of-the-art Lumina-DiMOO, it achieves over 4x acceleration of text-to-image generation while maintaining quality, significantly pushing the Pareto frontier of masked image generation.
% 
% \ifREVIEW
% \else
% The code and model weights are available \href{https://github.com/Kaiwen-Zhu/MIGM-Shortcut}{here}.
The code and model weights are available at \url{https://github.com/Kaiwen-Zhu/MIGM-Shortcut}.
% \fi
\end{abstract}

%% file: secs/1-intro-v2.tex
\section{Introduction}
\label{sec:intro}
Masked Image Generation Models (MIGMs)~\cite{maskgit,muse} present a prominent paradigm for visual generation. They typically model an image as a sequence of discrete tokens, and predict them from ``mask'' states progressively.
When scaled up, they have achieved strong performance that is comparable to the cutting-edge models~\cite{mimdatascaling, emigm}.
Moreover, their formulation can be seamlessly unified with the generation of other modalities. Recognizing this, the recently emerging multi-modal masked diffusion models have incorporated MIGMs' paradigm to exhibit versatile generation capabilities~\cite{mmada, dimoo}.

As MIGMs are increasingly integrated into large-scale foundation models, an imperative demand is to improve their efficiency, which is constrained by the multiple steps of bi-directional attention. One way is to reduce the number of generation steps. However, implementing this is hindered by the multi-modality problem~\cite{nat,remix}, that is, standard MIGMs struggle to model the joint distribution of multiple tokens in a single step.
Another way is to reduce the cost of each step. Inspired by the KV-cache mechanism in auto-regressive models, many works propose to reuse the features in previous steps considering that the token features often remain stable~\cite{recap, dllmcache, dimoo}. Despite some success, they are mostly based on intuitively handcrafted rules and the acceleration effect is limited.

\begin{figure}
    \centering
    % \subfloat[Latent trajectory.\label{fig:lat-traj}]{
    \includegraphics[width=.88\linewidth]{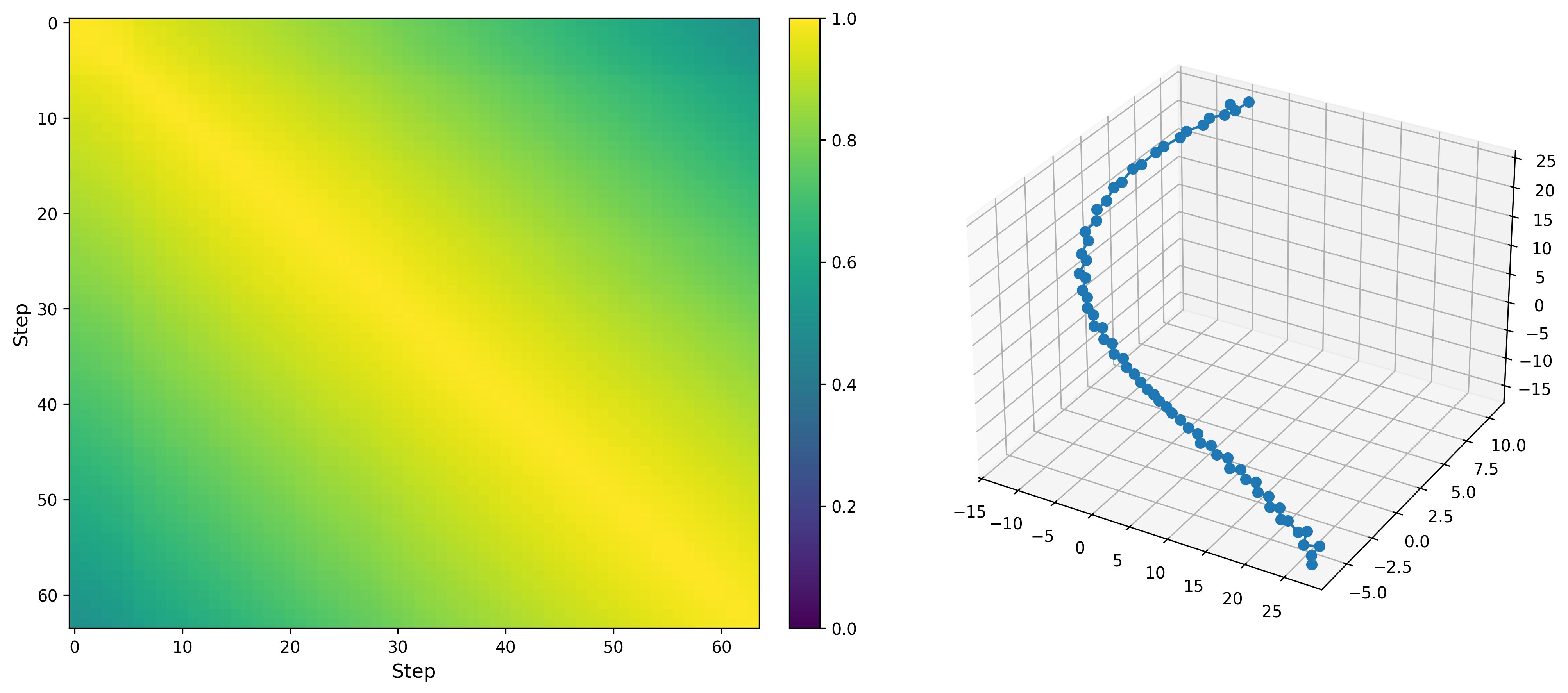}
    % }
    % \hfill
    % \subfloat[Token embedding trajectory.\label{fig:obs-traj}]{\includegraphics[width=.498\linewidth]{figs/traj/dynamics_token.png}}
    \caption{\textbf{Visualization of trajectory smoothness.} A point in the trajectory is the feature averaged over all tokens in a step. Left: heatmap of pairwise cosine similarity; Right: t-SNE visualization of the trajectory.}
    % \vspace{-0.5em}
    \label{fig:traj}
\end{figure}
\begin{figure}
    \centering
    % \subfloat[Images generated by MIGM.\label{fig:fork-img}]
    % {\includegraphics[width=.99\linewidth]{figs/fork_grid_flat.png}}
    % \\
    \subfloat[MIGM's feature trajectories.\label{fig:fork-traj}]
    {\includegraphics[width=.48\linewidth]{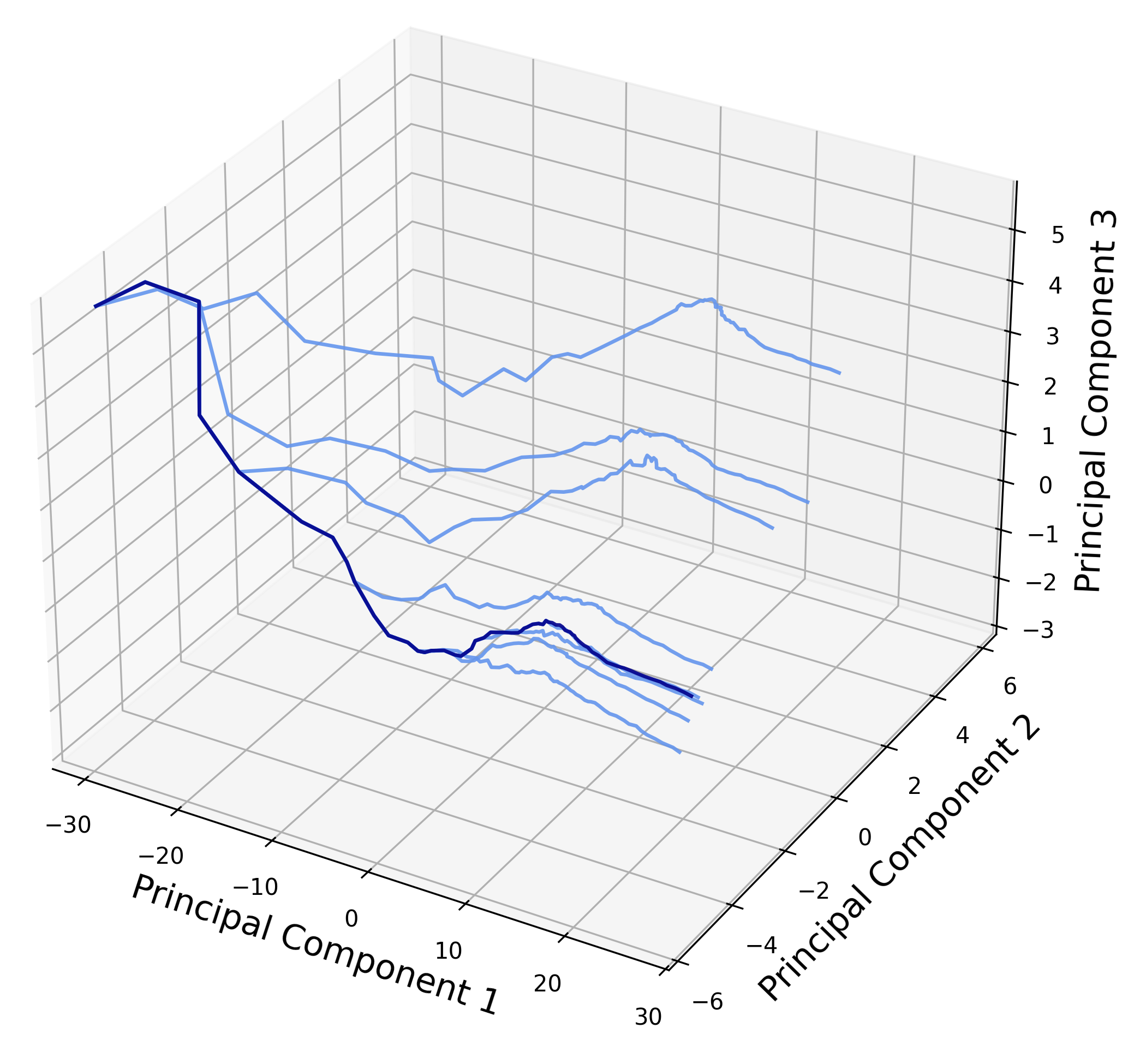}}
    \hfill
    \subfloat[Diffusion's feature trajectories using ODE sampling.\label{fig:fork-traj-ode}]
    {\includegraphics[width=.48\linewidth]{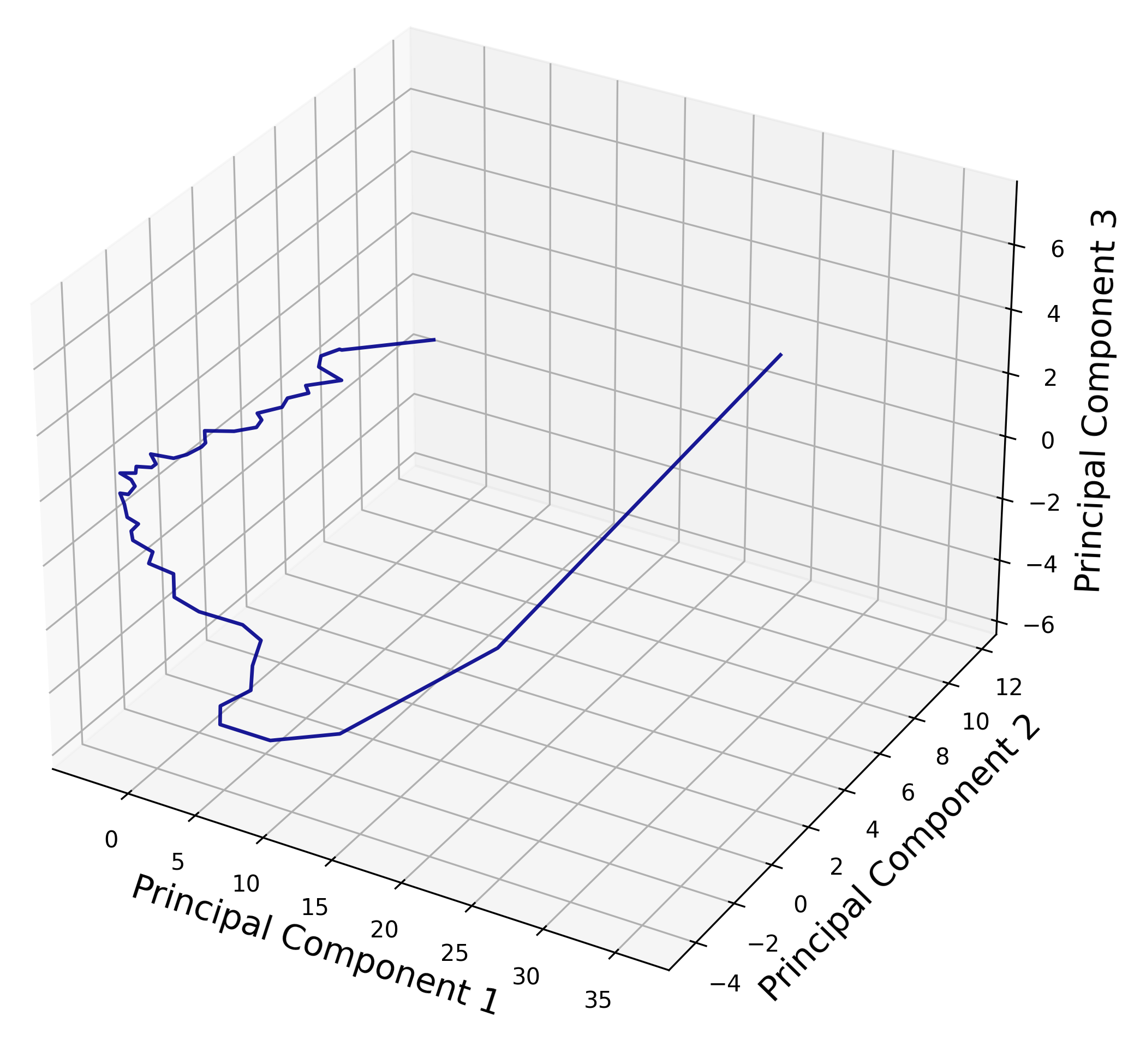}}
    % \vspace{-0.5em}
    \caption{\textbf{PCA visualization of feature trajectories generated with the same prompt and initial random seed}. (a) Using a MIGM, we first generate a trajectory (the dark one) and then change the random seed at intermediate steps to generate more samples (the light ones). The randomness in sampling tokens greatly affects the generation process. (b) In contrast, for continuous diffusion with ODE sampling, trajectories generated from the same starting point are always the same, without randomness at intermediate steps.}
    \label{fig:fork}
    % \vspace{-1.5em}
\end{figure}

The caching-based methods are also well-developed for continuous diffusion models, and a trend is to shift from reusing to predicting~\cite{cachesurvey}.
Some works view the features across steps as an evolving trajectory. They argue that this trajectory is smooth and thus can be predicted by simple computation. For example, TaylorSeer~\cite{taylorseer} and HiCache~\cite{hicache} predict the next feature by polynomial expansions, assuming the feature trajectory is sufficiently well-behaved.

Along this research line, we investigate whether the smoothness also holds true for MIGMs.
Using Lumina-DiMOO~\cite{dimoo} to generate samples, we investigate the features in the last layer of the model\footnote{In the rest of the paper, the term ``feature'' always refers to the last-layer feature. For the reason please refer to Appendix~\ref{supp:sec:feature-layer}.}. We compute the temporal self-similarity and visualize the trajectory using t-SNE~\cite{tsne}.
% Using Lumina-DiMOO~\cite{dimoo} to generate samples, we investigate the features in the last layer of the model\footnote{In the rest of the paper, the term ``feature'' always refers to the last-layer feature. For the reason please refer to Appendix~A.}. We compute the temporal self-similarity and visualize the trajectory using t-SNE~\cite{tsne}.

% 
% As a reference, we also do this for the trajectory of token embeddings. 
% 
An example is shown in \cref{fig:traj}. The features in consecutive steps are extremely similar, and the visualized trajectory exhibits a regular pattern.
% and the t-SNE visualization is not so jagged as that of the token embeddings.
% 
This suggests that as the model tries to learn the trajectory of token generation, it has also learned a latent trajectory in its internal feature space that underlies that generation process. This latent space may be more well-structured and thus more conducive to modeling.
% \footnote{Appendix \ref{supp:sec:lat} provides a digressive discussion on this.}.

In spite of the similar observation, we cannot directly apply the existing methods for continuous diffusion on MIGMs.
These methods require that the feature dynamics are self-contained, that is, the entire trajectory is uniquely determined by its own history, receiving no exogenous inputs. This is valid under the ODE sampling of continuous diffusion models.
However, it is not the case for MIGMs. As the starting point is always a fully masked sequence without randomness, the diversity must rely on the sampling in the generation process. The randomness in sampling tokens directs the trajectory to various samples.
As illustrated in \cref{fig:fork-traj}, starting from the same point, if we change the random seed at an intermediate stage of generation, then the trajectory will fork, generating diverse samples.
As a result, it is ill-posed to predict future features from the past features alone, which is what existing methods for continuous diffusion do. We must model the dynamics controlled by the observed sampling results~\cite{ncde}.

This paper seeks to tackle this challenge.
It greatly increases the modeling complexity to account for the sampled tokens, so the simple rule-based training-free methods~\cite{taylorseer, hicache, foca} are not optimal. We propose to learn the controlled latent dynamics with a neural network.
Based on the observation that the trajectory is smooth to some extent, this network may not require particularly high complexity.
Therefore, we choose to use a lightweight model to learn the feature dynamics. The model takes as input the feature and sampled tokens in the previous step, and predicts the direction to the next feature.
Once the feature dynamics are learned, at inference time we can replace the original heavy base model with this lightweight model in many steps, saving computation and accelerating generation.
We term this model as a \textit{shortcut}, since it takes the smooth path in the latent feature space to save computation.

We instantiate our method, \method, on two representative models, the seminal work MaskGIT~\cite{maskgit} and the recent state-of-the-art model Lumina-DiMOO~\cite{dimoo}. 
Extensive experiments validate the rationale and effectiveness of our method. For example, we achieve over $4\times$ speedup for Lumina-DiMOO with negligible performance drop on the text-to-image task. 
% 
% Interestingly, since a simple modification of the computation paradigm could bring such remarkable gains, we may need to rethink whether the current MIGM paradigm suffers from inherent limitations, such as information loss in sampling and computation redundancy.
% 
\method not only yields practical acceleration benefits, but also brings up a perspective of learning latent dynamics, which we demonstrate to be simple yet effective.
Beyond the acceleration effects, we hope \method could provide more insights into masked generative models and advance the development of this field.

%% file: secs/2-related_work.tex
\section{Related Work}
% \paragraph{Masked image generation.}
\subsection{Masked Image Generation}
Discrete image generation is initially dominated by auto-regressive models, which predict the next pixel or vector-quantized token following a specific order, usually as raster scan~\cite{pixelrnn, imagetransformer, zst2i, vqgan}.
They suffer from two limitations. First, the number of network evaluations equals the number of pixels or tokens, requiring prohibitive computation cost. Second, the causal attention restricts the model from attending to future tokens and thus limits the information interaction, but it is hard to predefine an appropriate generation order that could alleviate this limitation.
In view of this, MaskGIT~\cite{maskgit} proposes the masked image generation paradigm inspired by BERT~\cite{bert}. It takes as input a randomly masked token sequence and predicts multiple tokens within a single network evaluation, thereby overcoming inefficiency and the rigid generation order.
Moreover, courtesy of the flexibility in the conditioning masked positions, MIGMs can be applied to image editing without fine-tuning~\cite{maskgit, muse}.
It is also revealed that masked image modeling could help in representation learning~\cite{mae, mage, simmim, mimdatascaling}.
To further boost the performance of MIGMs, many works present improvements on various components, including the sampler~\cite{halton}, mask schedule~\cite{bagtricks}, hyper-parameter optimization~\cite{autonat}, and model architecture~\cite{enat, meissonic}, achieving performance comparable to the state-of-the-art models in image generation~\cite{emigm}.
It is interesting to note that MIGMs can be connected with masked diffusion, which performs the diffusion process on categorical data~\cite{d3pm, emigm}.
This paradigm enables natural unification of processing different modalities. Recognizing this, MMaDA~\cite{mmada} and Lumina-DiMOO~\cite{dimoo} formulate various tasks as modeling sequences interleaved with text and images, exhibiting versatile capabilities in multi-modal generation and understanding.

% \paragraph{Image generation acceleration.}
\subsection{Image Generation Acceleration}
High-quality image generation often requires multiple steps, incurring significant latency.
To alleviate this, many works propose to reduce the number of steps, achieving few-step or even one-step generation for continuous diffusion models~\cite{progressivedistill, dpmsolver, consistencymodels, dmd, continuousshortcut, meanflow, flashdmd}. 
There are also a few works making such efforts for discrete diffusion models~\cite{dimo, klass}. Despite success to some extent, they struggle to sufficiently address the multi-modality problem~\cite{nat}.
Other works seek to reduce the cost of each generation step. Typical approaches include pruning~\cite{diffpruning}, quantization~\cite{diffquantization}, and caching~\cite{cachesurvey}. Our work is most closely related to caching-based methods.
They observe that the learned token features in consecutive steps are often highly similar, so it is possible to approximate the features using those in previous steps~\cite{deepcache, fora, skipdit}. To reduce the approximation error, many methods have been developed to adaptively identify reusable features~\cite{deltadit, lazymar, dimoo} or improve the refresh schedule~\cite{eoc, blockcaching, teacache}.
Beyond directly replacing target features with cached features, some works perform cheap transformations on them. For example, FoCa~\cite{foca}, TaylorSeer~\cite{taylorseer}, and HiCache~\cite{hicache} treat the feature as a function of time, so future features can be predicted using ODE solvers or polynomial expansions. However, these methods often require the feature to be differentiable with respect to time, which does not hold true for discrete diffusion.
TeaCache~\cite{teacache} and Block Caching~\cite{blockcaching} represent target features as polynomials of cached features and learn the coefficients, as a trick to refine the cache.
In contrast to these works, our method focuses on learning a feasible feature dynamics model with moderate complexity, and can be readily applied to various existing well-trained base models.

% \subsection{Learning Latent Representations and Dynamics}
% Obtaining the effective representation of data is a long-standing goal of machine learning. Many researchers advocate that modeling the latent representation offers a particularly promising pathway towards intelligence~\cite{worldmodel, jepa, ijepa, ldm, rae, latentreasonsurvey, repa}. Furthermore, for sequence data such as time series, the problem extends to modeling the dynamics of latent representations~\cite{vrnn, dvbf, planet}.
% % 

% \todo{}

%% file: secs/3-method.tex
\section{Method}
We first define the notations and formulate the shortcut model (\cref{sec:formulation}). Next, we provide intuition and empirical evidence for our assumptions (\cref{sec:rationale}). Then we outline the concrete implementation of the neural network (\cref{sec:model}). Finally, we describe the training (\cref{sec:train}) and usage (\cref{sec:infer}) of the model.
\subsection{Formulation}
\label{sec:formulation}
\paragraph{Preliminary and Notation.} 
We adopt the typical framework of discrete masked image generation~\cite{maskgit, muse, mage, meissonic}. Let $\mathcal{V}$ denote the vocabulary of the discrete tokenizer. A special token $\mask$ is introduced to form the augmented vocabulary $\widetilde{\mathcal{V}}=\mathcal{V}\cup\{\mask\}$.
The generation process is represented by $\{\boldsymbol{x}_t\}_{t\in[0,1]}$, where $\boldsymbol{x}_t \in \widetilde{\mathcal{V}}^L$ and $L$ is the number of tokens in the image. Starting from the fully masked $\boldsymbol{x}_0 = (\mask, \mask, \cdots,\mask)$, the number of masked tokens progressively reduces, leading to the fully unmasked image $\boldsymbol{x}_1 \in \mathcal{V}^L$.
Formally, define a scheduler $\gamma: [0,1] \rightarrow [0,1]$ so that $\gamma(0)=1, \gamma(1)=0$, and $\gamma$ monotonically decreases. $\gamma(t)$ represents the mask ratio at time $t$.
In practice, the generation process is discretized into $N$ steps $\{t_i=i/N\}_{i=0}^N$. At step $i$, the unmasking model $M$ predicts the distribution over $\mathcal{V}$ of each token $\boldsymbol{p}_{t_i}=M(\boldsymbol{x}_{t_{i-1}})$.
Sampling from $\boldsymbol{p}_{t_i}$ yields $\hat{\boldsymbol{x}} \in \mathcal{V}^L$. With some strategy, $N_{t_i}=\lceil L\cdot\gamma(t_{i-1})\rceil - \lceil L\cdot\gamma(t_i)\rceil$ masked positions in $\boldsymbol{x}_{t_{i-1}}$ are selected and filled with the corresponding tokens in $\hat{\boldsymbol{x}}$, obtaining $\boldsymbol{x}_{t_i}$. A typical selection strategy is the confidence-based sampler~\cite{maskgit}.
Denote this sampling process by $\boldsymbol{x}_{t_i} \sim K(\cdot | \boldsymbol{x}_{t_{i-1}}, \boldsymbol{p}_{t_i}, N_{t_i})$.
The above procedure repeats for $i=1,2,\cdots, N$ to finalize the sample $\boldsymbol{x}_1$.
\par

\paragraph{Model.} 
Let $\boldsymbol{f}_{t_i} = M_f(\boldsymbol{x}_{t_{i-1}}) \in \mathbb{R}^{L\times D}$ denote the last hidden state of the unmasking model $M$, that is, $\boldsymbol{p}_{t_i} = \text{softmax}(H(\boldsymbol{f}_{t_i}))$, where $H$ is the classification head.
We view $\{\boldsymbol{f}_{t_i}\}_{i=1}^N$ as a controlled latent trajectory that underlies the generation process, and propose to capture its dynamics by learning.
Treating $\{\boldsymbol{x}_{t}\}$ as the observation and $\{\boldsymbol{f}_{t}\}$ as the state, we reformulate the generation process as a state-space model:
% \par
\\
State transition
\begin{equation} 
\label{eq:trans}
\boldsymbol{f}_{t_{i+1}} = \boldsymbol{f}_{t_{i}} + S_\theta(\boldsymbol{f}_{t_{i}}, \boldsymbol{x}_{t_{i}}, t_{i}) + \boldsymbol{\epsilon},
\end{equation}
% \par
Observation
\begin{equation} 
\label{eq:obs}
\boldsymbol{x}_{t_{i+1}} \sim K(\cdot | \boldsymbol{x}_{t_{i}}, \text{softmax}(H(\boldsymbol{f}_{t_{i+1}})), N_{t_{i+1}}).
\end{equation}
Here, $S_\theta$ is a lightweight neural network with learnable parameter $\theta$, and $\boldsymbol{\epsilon}$ is the error term that follows a normal distribution with zero mean.
We call $S_\theta$ the \emph{shortcut} model, highlighting that it bypasses the cumbersome base model $M$ by taking a shortcut from $\boldsymbol{f}_{t_{i}}$ that directly leads to $\boldsymbol{f}_{t_{i+1}}$.

\begin{figure}
    \centering
    \subfloat[Scatter plot of the norms of input/target difference.\label{fig:lip-scatter}]
    {\includegraphics[width=.493\linewidth]{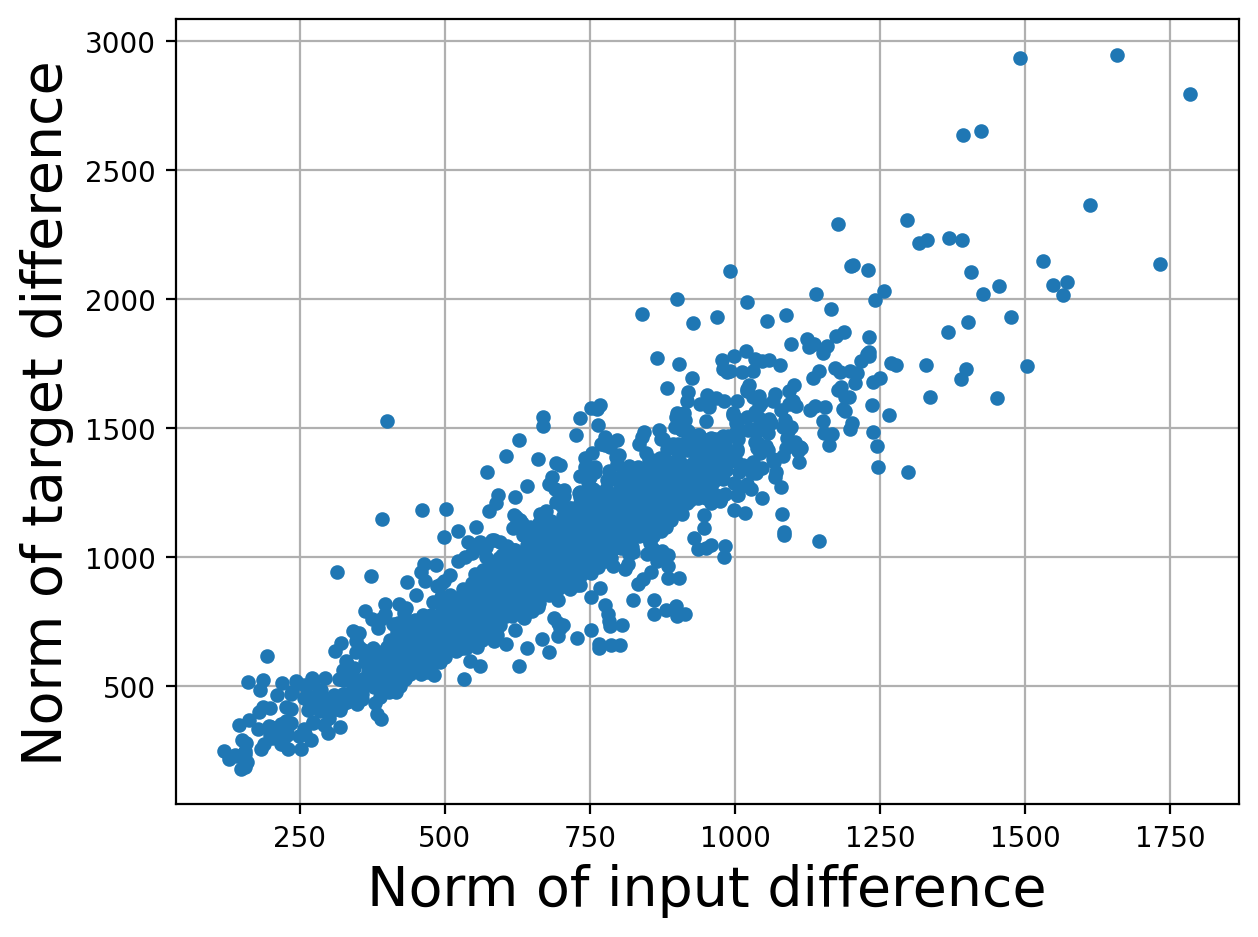}}
    \hfill
    \subfloat[Frequency histogram of the ratio of the two norms.\label{fig:lip-hist}]
    {\includegraphics[width=.493\linewidth]{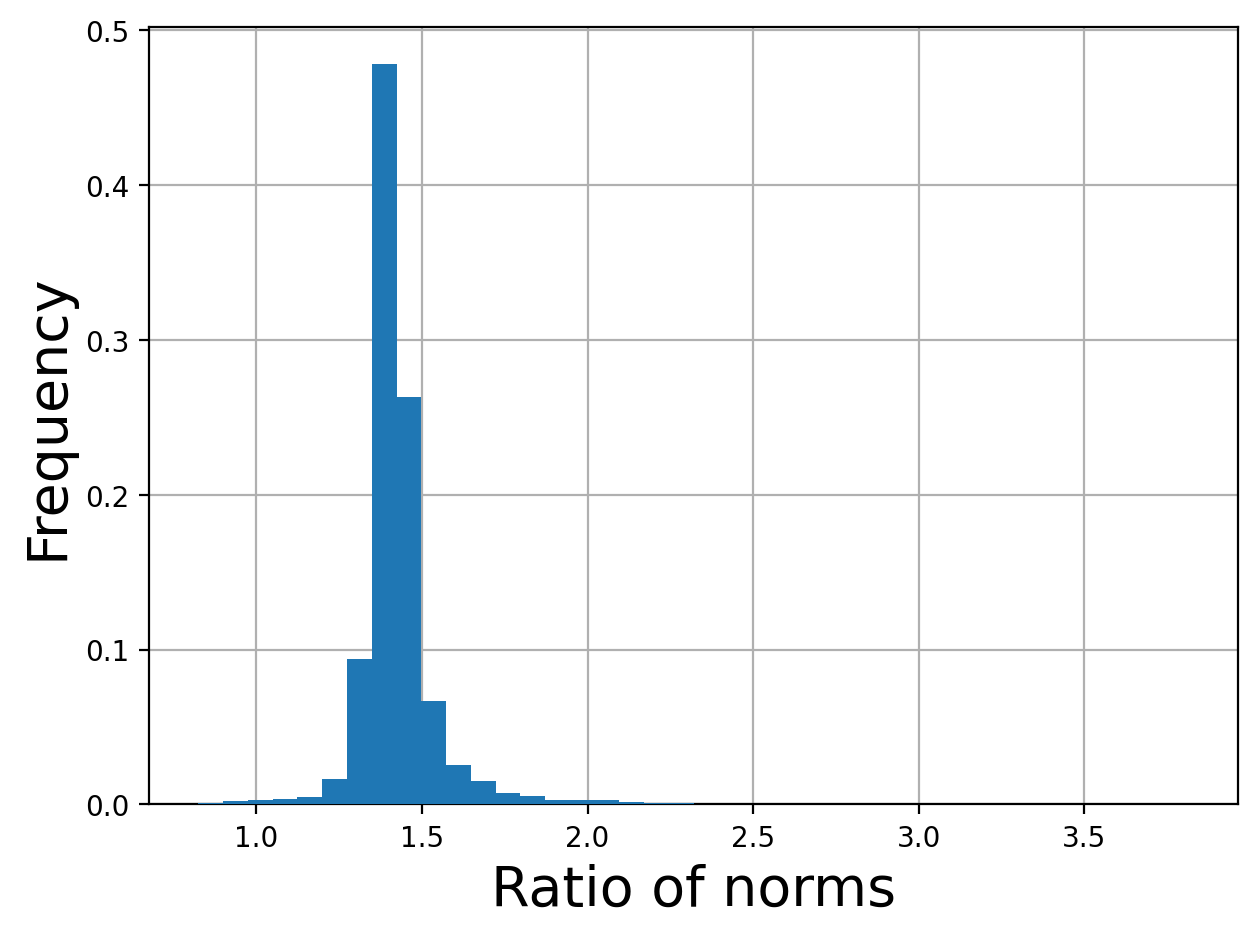}}
    \caption{\textbf{Observation of local Lipschitz behavior of map \eqref{eq:map}.} For neighboring points along trajectories, we compute the norms of the target and input differences. The ratio of the two norms concentrates around a moderate constant.}
    \label{fig:lip}
\end{figure}

\subsection{Rationale} 
\label{sec:rationale}
In this section, we discuss why the map\footnote{The time $t_i$ is omitted in the input for brevity. In fact, it is more like an auxiliary term and can be implicitly captured by $\boldsymbol{f}_{t_i}$~\cite{zheng2025masked}.} 
\begin{equation}
\label{eq:map}
(\boldsymbol{f}_{t_{i}},\boldsymbol{x}_{t_{i}}) \mapsto \boldsymbol{f}_{t_{i+1}} - \boldsymbol{f}_{t_i}
\end{equation}
of the shortcut model $S_\theta$ could have far lower complexity than the map $\boldsymbol{x}_{t_{i}} \mapsto \boldsymbol{f}_{t_{i+1}}$ of the original base model $M$.
As \cref{fig:traj} shows, $\boldsymbol{f}_{t_{i}}$ and $\boldsymbol{f}_{t_{i+1}}$ are very close, with a cosine similarity usually larger than 0.95.
Note that such high similarity is not merely because they share the same space. In fact, in the sampling process \eqref{eq:obs}, many tokens collapse into $\mask$, discarding the rich information contained in the continuous features. This information loss is not trivially reversible. 
As a result, to obtain $\boldsymbol{f}_{t_{i+1}}$ solely from $\boldsymbol{x}_{t_{i}}$, the base model $M$ must compute the features of masked tokens from scratch, which is already largely contained in $\boldsymbol{f}_{t_{i}}$. Therefore, taking $\boldsymbol{f}_{t_{i}}$ as input to supplement information will significantly alleviate the computational burden.
However, using $\boldsymbol{f}_{t_{i}}$ alone is insufficient either, as the sampling process \eqref{eq:obs} itself also contains crucial information. The sampled $\boldsymbol{x}_{t_{i}}$ directly points out the direction from $\boldsymbol{f}_{t_{i}}=M_f(\boldsymbol{x}_{t_{i-1}})$ to $\boldsymbol{f}_{t_{i+1}}=M_f(\boldsymbol{x}_{t_{i}})$.
Without $\boldsymbol{x}_{t_{i}}$, the map would be stochastic, hence when trained in a supervised learning manner, the model inevitably outputs blurry expectations~\cite{pix2pix}.
Collectively, combined with the residual connection~\cite{resnet} in \cref{eq:trans}, the map \eqref{eq:map} describes how the latent state evolves driven by a newly sampled observation.
In summary, the computation of this map is self-contained and efficient, thanks to the smoothness of the latent dynamics.

To support the simplicity of the map \eqref{eq:map}, we inspect the local regularity on observed data.
We collect 100 trajectories $\{(\boldsymbol{x}_{t_i}^{(k)},\boldsymbol{f}_{t_i}^{(k)}), i=1,\cdots,N\}_{k=1}^{100}$ using Lumina-DiMOO~\cite{dimoo}. Let $\boldsymbol{u}_i^{(k)}$ denote the concatenation of $E(\boldsymbol{x}_{t_i}^{(k)})$ and $\boldsymbol{f}_{t_i}^{(k)}$, where $E$ is the token embedding, and let $\boldsymbol{\Delta}_{i}^{(k)}$ denote $\boldsymbol{f}_{t_{i+1}}^{(k)} - \boldsymbol{f}_{t_i}^{(k)}$. That is, $\boldsymbol{u}_i^{(k)}$ and $\boldsymbol{\Delta}_{i}^{(k)}$ represent the input and target of map \eqref{eq:map}, respectively.
We compute the Frobenius norm of their finite differences $\lVert \boldsymbol{u}_{i+1}^{(k)} - \boldsymbol{u}_i^{(k)}\rVert_F$ and $\lVert \boldsymbol{\Delta}_{i+1}^{(k)} - \boldsymbol{\Delta}_{i}^{(k)} \rVert_F$, and present the scatter plot and frequency histogram of their ratio in \cref{fig:lip}.
It is observed that the points $\left(\lVert \boldsymbol{u}_{i+1}^{(k)} - \boldsymbol{u}_i^{(k)}\rVert_F, \lVert \boldsymbol{\Delta}_{i+1}^{(k)} - \boldsymbol{\Delta}_{i}^{(k)} \rVert_F\right)$ mostly lie close to a straight line passing through the origin; in other words, the ratio $\lVert \boldsymbol{\Delta}_{i+1}^{(k)} - \boldsymbol{\Delta}_{i}^{(k)} \rVert_F / \lVert \boldsymbol{u}_{i+1}^{(k)} - \boldsymbol{u}_i^{(k)}\rVert_F$ concentrates around a constant.
This observation suggests an approximately uniform local Lipschitz behavior rather than highly variable local amplification. Under this assumption, we can anticipate that the model $S_\theta$ is not necessarily required to have much high-frequency representational capacity.

% \begin{figure}
%     \centering
%     \includegraphics[width=0.7\linewidth]{figs/delta_pca.png}
%     \caption{\textbf{Cumulative explained variance ratio (EVR) of PCA on feature updates.} Around 10\% principal components contribute 99\% variance.}
%     \label{fig:update-pca}
% \end{figure}

\subsection{Shortcut Model}
\label{sec:model}
The guideline for designing the shortcut model is to be simple and lightweight.
The backbone consists of only a cross-attention layer followed by a self-attention layer~\cite{transformer}. The cross-attention is responsible for absorbing the information from the sampled tokens, and the self-attention could transform the collected information into the evolving direction.
In practice, besides the features, what is fed into the model is the tokens newly decoded in the previous step, carrying position embeddings, instead of the full token sequence. These tokens function as the source of key and value for the cross-attention. This highlights that the evolution of features is sufficiently driven by the new tokens, and there is also a practical benefit of efficiency.
To further save computational cost, we impose a bottleneck on the model. The inputs are projected to a lower-dimensional space, and projected back at the end of the model. Each projection is implemented by a linear layer. 
This is reasonable based on the assumption that the latent trajectory evolution is driven by a few new tokens and thus relatively low-rank.
% 
% To validate it, we randomly generate one hundred 64-step trajectories of Lumina-DiMOO and compute the temporal differences of features averaged over tokens in a step, totaling 6300 data points. Each data point has a dimension of 4096. We conduct PCA on them and plot the cumulative explained variance versus the number of principal components in \cref{fig:update-pca}.
% % 
% The rapid saturation may suggest that the feature updates concentrate in a low-dimensional subspace. Therefore, a model with lower hidden dimensions is still likely to have enough capacity to process the features.

Additionally, we include time as a conditioning input. The time scalar is transformed into sinusoidal embeddings~\cite{ddpm} and then modulates the features via adaptive layer normalization~\cite{dit}.
We empirically find that it could accelerate convergence. This may be because the dynamics are prominently time-variant, so the time conditioning helps the model to be aware of the current stage and accordingly steer its prediction~\cite{meissonic}.

\subsection{Training}
\label{sec:train}
The training of the shortcut model is straightforward. We collect a dataset of $(\boldsymbol{f}_{t_{i}}, \boldsymbol{x}_{t_{i}}, t_i, \boldsymbol{f}_{t_{i+1}})$ as training samples, and optimize the Mean Squared Error (MSE) loss
\begin{equation}
\label{eq:mse}
\mathbb{E}\left[\left\Vert\boldsymbol{f}_{t_{i+1}} - \left(\boldsymbol{f}_{t_{i}} + S_\theta\left(\boldsymbol{f}_{t_{i}}, \boldsymbol{x}_{t_{i}}, t_i\right)\right)\right\Vert_2^2\right].
\end{equation}
% 
% This is equivalent to maximizing the likelihood since the error term in the model \eqref{eq:trans} is assumed to follow a zero-mean normal distribution.
% 
Only the shortcut model's parameter $\theta$ is trainable, and the base model remains frozen.
Under the modeling of \eqref{eq:trans}, optimizing this objective maximizes the likelihood.
% 
% The form of the objective resembles that in knowledge distillation~\cite{fitnet, selfdistill}. The subtle difference is that the feature to regress is not auxiliary implicit knowledge, but the only explicit target, and the input is not the same as the teacher's input, but the previous step's feature derived by our insights for masked generation. Further discussion is in Appendix~H.
The form of the objective resembles that in knowledge distillation~\cite{fitnet, selfdistill}. The subtle difference is that the feature to regress is not auxiliary implicit knowledge, but the only explicit target, and the input is not the same as the teacher's input, but the previous step's feature derived by our insights for masked generation. Further discussion is in Appendix~\ref{supp:sec:related_work}.

We have also explored more sophisticated training strategies. We tried to supplement the loss with a distribution matching term like Kullback-Leibler divergence as a regularizer. We also tried to expose the shortcut model to its own predictions as input~\cite{sfd, sf}.
However, it turns out that the simple MSE loss \eqref{eq:mse} alone works as well as these modifications. This in turn partially supports our assumption that the dynamics are easy enough to be learned by a simple approach, whereas complex designs bring marginal improvements.
% 
% More details are in Appendix~B.
More details are in Appendix~\ref{supp:sec:attempts}.

\begin{figure}
    \centering
    \includegraphics[width=\linewidth]{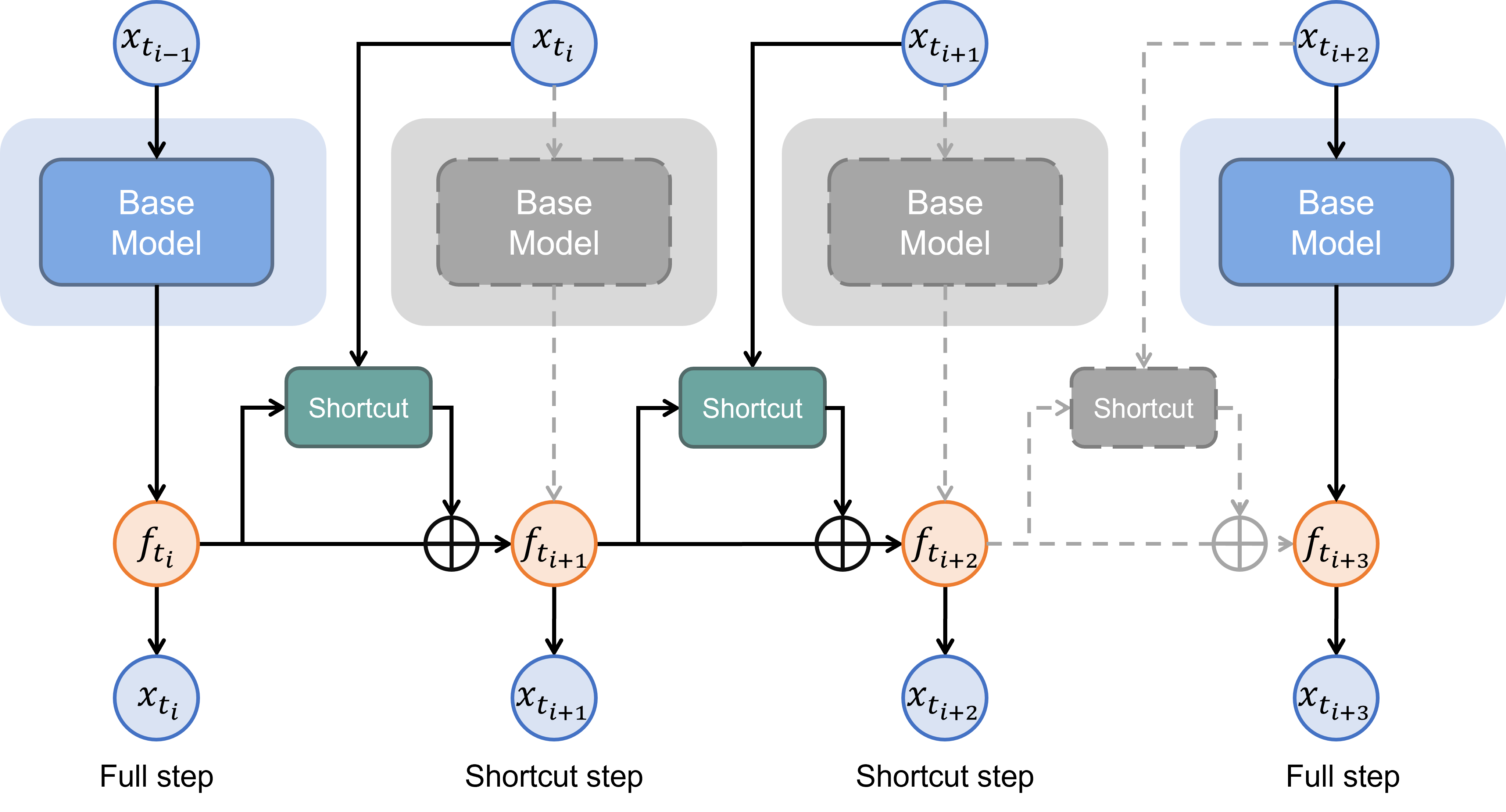}
    \caption{\textbf{Inference workflow of \method.} Colored blocks and solid lines represent activated computation, while gray blocks and dash lines represent suppressed computation. At a full step, the inference is the same as the vanilla procedure, taking $\boldsymbol{x}_{t_{k-1}}$ as input to compute $\boldsymbol{f}_{t_k}$ using the base model. At a shortcut step, the shortcut model takes both $\boldsymbol{x}_{t_{k-1}}$ and $\boldsymbol{f}_{t_{k-1}}$ as input to compute $\boldsymbol{f}_{t_k}$, skipping the base model.}
    \label{fig:pipeline}
\end{figure}

\subsection{Inference}
\label{sec:infer}

At inference time, we first compute $\boldsymbol{f}_{t_1}$ using the base model $M$ from the masked sequence $\boldsymbol{x}_0$, and in subsequent steps, we can replace the heavy computation of $\boldsymbol{f}_{t_{i+1}} = M_f(\boldsymbol{x}_{t_{i}})$ with the light $$\hat{\boldsymbol{f}}_{t_{i+1}} = \boldsymbol{f}_{t_{i}} + S_\theta(\boldsymbol{f}_{t_{i}},\boldsymbol{x}_{t_{i}},t_i).$$
However, in fact $\hat{\boldsymbol{f}}_{t_{i+1}}$ does not equal $\boldsymbol{f}_{t_{i+1}}$ due to the error term in \eqref{eq:trans} and the optimization error, so continuously applying the shortcut model will cause distribution shift and keep amplifying the error.
To suppress error accumulation, we regularly apply the base model to obtain the feature that follows the correct distribution.
Specifically, suppose we generate the image in $N$ steps and have a budget to invoke the base model for $B$ times, then at step $1+\lfloor jN / B\rfloor, j=0,1,\cdots,B-1$, we invoke the base model to slowly but accurately compute the feature (full step), and at other steps we invoke the shortcut model to approximately but rapidly compute the feature (shortcut step).
\cref{fig:pipeline} provides an overview of this procedure.

%% file: secs/4-exp.tex
\section{Experiments}
\begin{table}
    \centering
    \caption{\textbf{Performance of MaskGIT with and without shortcut.} The speedup is with respect to 15-step vanilla, which is the recommended setting of \citet{maskgittorch}. The best and second-best performances are marked in \textbf{bold} and \underline{underline}.}
    \resizebox{\linewidth}{!}{
    \input{tabs/maskgit}
    }
    \label{tab:maskgit}
\end{table}

In this section, we validate the soundness of \method by extensive experiments.
Implementation of attention modules and inference hyper-parameters (temperature, mask schedule, classifier-free guidance scale, etc.) are the same as the base MIGM. The bottleneck ratio is 2. All experiments are conducted on H200 GPUs.

As a proof of concept, we first apply our method to the seminal MIGM MaskGIT~\cite{maskgit} to probe the effectiveness (\cref{sec:maskgit}). 
Then we perform experiments on Lumina-DiMOO~\cite{dimoo}, a multi-modal masked diffusion language model that achieves state-of-the-art text-to-image capability. We will show that equipping it with \method, the generation can be greatly accelerated while maintaining the image quality (\cref{sec:dimoo}).
We also ablate key designs of \method to further support our motivation (\cref{sec:ablation}).

\begin{figure*}
\begin{minipage}{0.29\linewidth}
\includegraphics[width=\linewidth]{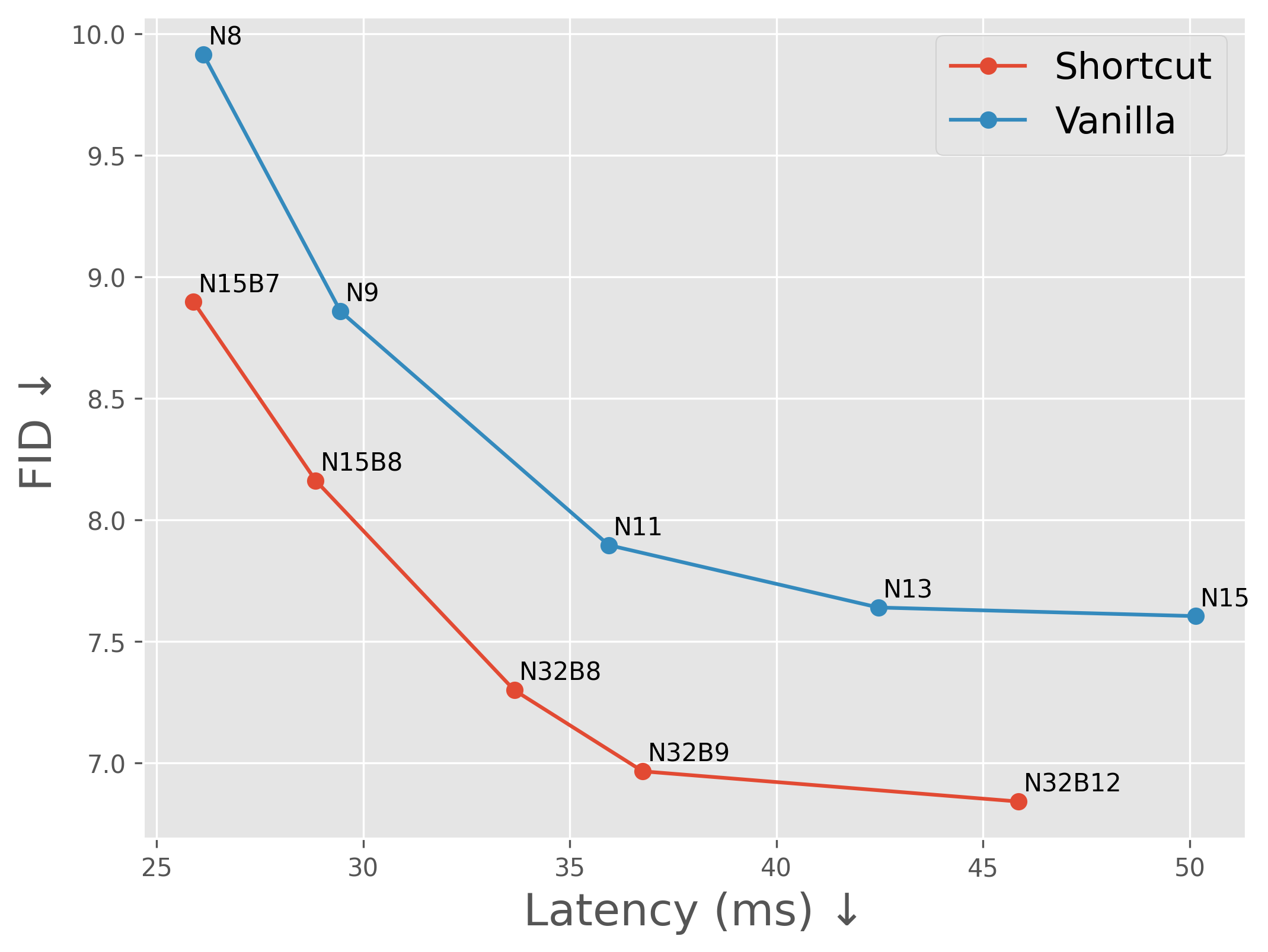}
\captionof{figure}{\textbf{Quality-speed trade-off MaskGIT with and without shortcut.} MaskGIT-Shortcut can reach lower FID and faster speed. The number of steps $N$ and the budget $B$ are marked. \label{fig:maskgit}}
\end{minipage}
\hfill
\begin{minipage}{0.68\linewidth}
\vspace{-1em}
\includegraphics[width=\linewidth]{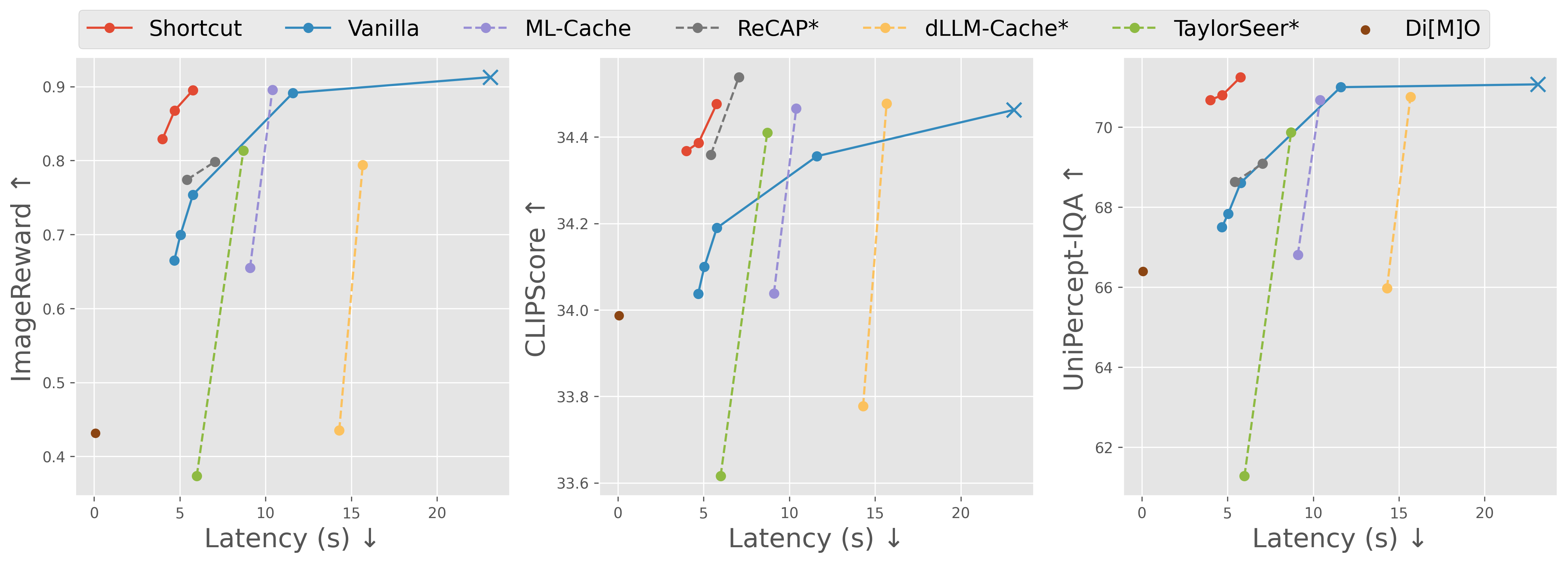}
\captionof{figure}{\textbf{Quality-speed trade-off of different methods accelerating text-to-image generation using Lumina-DiMOO.} The point before acceleration is denoted by $\times$. Dash lines denote existing acceleration methods.
% DiMOO-Shortcut (red line) significantly pushes the Pareto frontier. 
\label{fig:main}}
\end{minipage}
\vspace{-1em}
\end{figure*}

\begin{figure*}
    \centering
    \includegraphics[width=.95\linewidth]{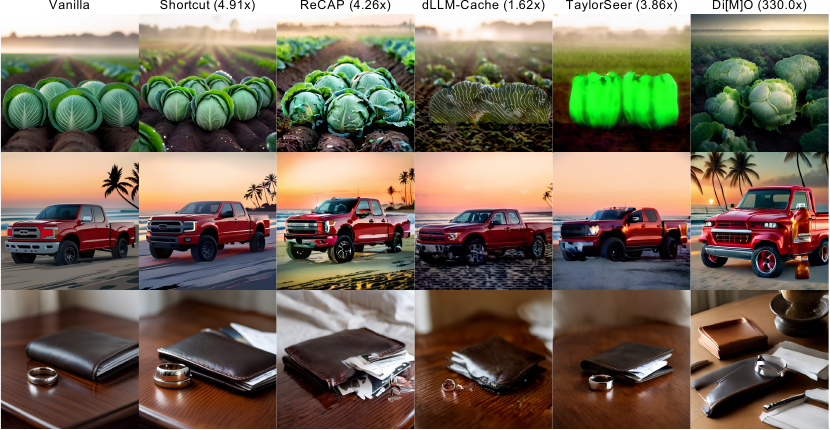}
    \caption{\textbf{Qualitative comparison of acceleration methods.} DiMOO-Shortcut achieves decent quality with $4.9\times$ speedup.}
    \label{fig:qual}
    \vspace{-0.5em}
\end{figure*}

\begin{table*}
    \centering
    \caption{\textbf{Performance of different methods accelerating text-to-image generation using Lumina-DiMOO, along with the one-step model \dimo.} ``Memo.'' denotes peak GPU memory. Apart from the performance before acceleration (Vanilla 64 steps), for each metric the best and second-best entries are marked in \textbf{bold} and \underline{underline}, respectively. For ML-Cache, $w,r,c$ represent \texttt{warmup\_ratio}, \texttt{refresh\_interval}, and \texttt{cache\_ratio} in their paper. * denotes methods re-implemented by us on Lumina-DiMOO. To emphasize the effects of acceleration methods, the record of latency and memory excludes the stage of decoding tokens.}
    \label{tab:main}
    % \resizebox{1.008\linewidth}{!}{
    \input{tabs/main}
    % }
\end{table*}

\begin{figure*}
\begin{minipage}{0.32\linewidth}
\includegraphics[width=\linewidth]{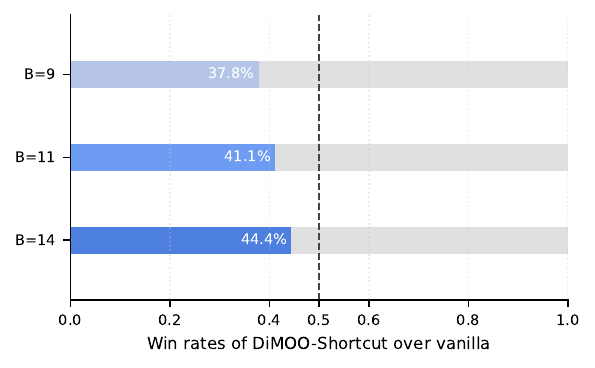}
\captionof{figure}{\textbf{Win rates of DiMOO-Shortcut over vanilla in the human study.} DiMOO-Shortcut emulates the vanilla model in perceptual quality in most cases. \label{fig:humanstudy}}
\end{minipage}
\hfill
\begin{minipage}{0.66\linewidth}
\includegraphics[width=\linewidth]{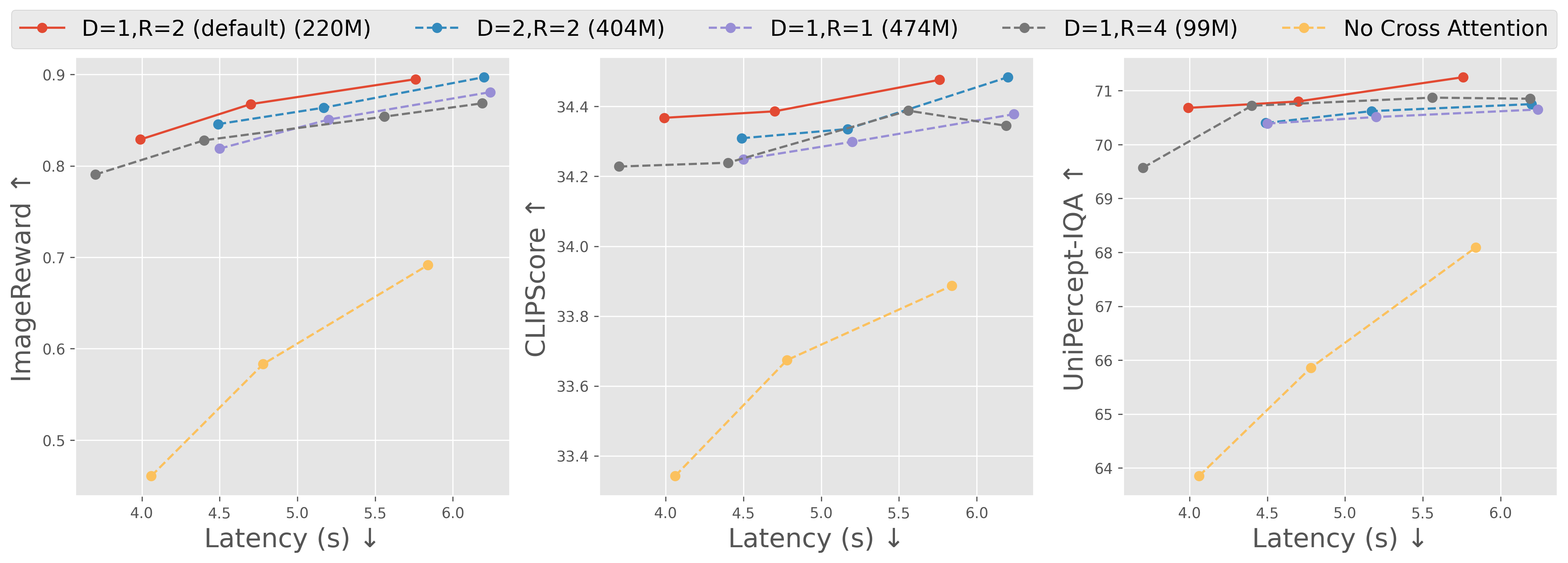}
\vspace{-2em}
\captionof{figure}{\textbf{Quality-speed tradeoff of different model designs of DiMOO-Shortcut.} The solid red line denotes the default, and the dash lines denote ablated variants. All models infer with 64 steps and budget 14/11/9, and the lighter variant $D=1,R=4$ has an extra budget 16.\label{fig:ablation}}
\end{minipage}
\end{figure*}

\subsection{Class to Image using MaskGIT}
\label{sec:maskgit}
The MaskGIT model we use is from \citet{maskgittorch}. Built upon this, we construct MaskGIT-Shortcut with 8.6M parameters, about $1/20$ of the base model. The latency of one invocation is reduced by $24\times$.
We use MaskGIT to generate 50,000 images of resolution $512\times 512$ from the class labels of ImageNet~\cite{imagenet} in 15 steps. We also use classifier-free guidance, so each image contributes $(15-1)\times2=28$ training samples, for a total of 1.4M.
We train MaskGIT-Shortcut for 5 hours using 4 H200 GPUs.

We evaluate the generation quality on ImageNet-512 by FID~\cite{fid}. The performance of MaskGIT with and without shortcut under different configurations is shown in \cref{tab:maskgit} and \cref{fig:maskgit}.
We can see MaskGIT-Shortcut can consistently generate better images at a faster speed.
By increasing the total steps, MaskGIT-Shortcut could even surpass the vanilla's optimal performance.
% 
% This phenomenon is somewhat strange. Intuitively, the upper bound of MaskGIT-Shortcut performance should have been the vanilla MaskGIT with the same number of steps. However, in our experiments, the MaskGIT-Shortcut with 32 steps and budget 8/9/12 all surpass vanilla MaskGIT with 32 steps.
% % 
% We provide a possible explanation as follows. It has been observed that increasing the number of steps may degrade the performance~\cite{maskgit}. For example, \citet{maskgittorch} reports the optimal number of steps as 15.
% % 
% That is to say, the feature trajectories under 15 steps are better than those under 32 steps.
% % 
% Our MaskGIT-Shortcut is trained on feature trajectories generated by MaskGIT with 15 steps. If well-trained, it will evolve along this golden trajectory that more accurately leads to the target data distribution than the 32-step trajectory, thus achieving lower FID.
% % 
% Besides, when compared with vanilla MaskGIT with 15 steps, MaskGIT-Shortcut with 32 steps takes the same trajectory but moves with a finer step size, thus also able to achieve better performance.
% % 
% This hypothesis highlights that as \method learns the latent dynamics, it may improve not only efficiency but also performance.

\subsection{Text to Image using Lumina-DiMOO}
\label{sec:dimoo}

\paragraph{Setup.}
The shortcut model applied to Lumina-DiMOO is termed DiMOO-Shortcut. It has 220M parameters, only $1/37$ of the original Lumina-DiMOO. The latency of one invocation is around $1/30$ of the base model.
We use prompts from \citet{blip3o} to prompt Lumina-DiMOO to generate 2000 images of resolution $1024\times1024$ in 64 steps. Each image corresponds to $(64-1) \times 2=126$ training samples, totaling 252K.
We train DiMOO-Shortcut for 12 hours using 4 H200 GPUs, which is also a modest cost.

\paragraph{Evaluation protocol.}
Following common practice~\cite{foca,taylorseer,hicache}, we evaluate the quality of generated images by ImageReward~\cite{imagereward} and CLIPScore~\cite{clipscore}. We also report UniPercept-IQA~\cite{unipercept}, a recent advanced metric based on multi-modal large language models. Peak GPU memory is also reported. We use the 1065 prompts from \citet{dpg}, and the resolution is $1024\times1024$.

\paragraph{Compared methods.}
We compare DiMOO-Shortcut with various acceleration methods. The most straightforward baseline is reducing the steps of Lumina-DiMOO. It also has a dedicated training-free acceleration technique called ML-Cache, which reports $2\times$ speedup.
Other compared methods include ReCAP~\cite{recap}, dLLM-Cache~\cite{dllmcache}, and TaylorSeer~\cite{taylorseer}. They are all training-free but not officially implemented on Lumina-DiMOO, so we adapt them to it.
ReCAP~\cite{recap} is a general acceleration method for MIGMs. dLLM-Cache~\cite{dllmcache} is designed for diffusion language models, but the open-source code repository also implements it on text-to-image using MMaDA~\cite{mmada}, which shares a similar architecture with Lumina-DiMOO. TaylorSeer~\cite{taylorseer} is a cutting-edge method for continuous diffusion, forecasting the next feature from previous features.
Besides these generally applied techniques, we also include a distilled one-step MIGM \dimo~\cite{dimo} for comparison.
% , which is distilled from Meissonic~\cite{meissonic}.

\paragraph{Accelerating effects.}
All the methods except \dimo have several configurable hyper-parameters to allow for a flexible quality-speed tradeoff. We carefully tune them and report the Pareto-optimal configurations of each method.
For each of these methods, we report two configurations: one that barely maintains quality while maximizing speed, and one that becomes a little faster but considerably compromises the quality.
The acceleration is with respect to Lumina-DiMOO using 64 steps.
\cref{tab:main} lists the results, and \cref{fig:main} provides a more clear visualization.
We can see that all three configurations of DiMOO-Shortcut approach the performance of the vanilla, while reaching $4.0\sim 5.8$ acceleration rate.
Notably, under budget $B=14$, DiMOO-Shortcut achieves the highest ImageReward and UniPercept-IQA, and the second highest CLIPScore, with a high acceleration rate of 4. Although ReCAP can achieve a slightly higher CLIPScore, its two other metrics are very low, and the acceleration rate is also lower.
% 
% Appendix~C evaluates on more benchmarks and generalization ability.
Appendix~\ref{supp:sec:eval} evaluates on more benchmarks and generalization ability.
% 
% Some qualitative examples are shown in \cref{fig:qual}, also supporting the superiority of DiMOO-Shortcut. For more examples, please refer to Appendix~D.
Some qualitative examples are shown in \cref{fig:qual}, also supporting the superiority of DiMOO-Shortcut. For more examples, please refer to Appendix~\ref{supp:sec:more_visualization}.
Note that although DiMOO-Shortcut consumes more memory, the extra cost is modest compared with the caching-based methods. This is because the cache contains high-dimensional tensors and needs to be maintained across steps, while our method only needs to load a lightweight shortcut model.

\begin{figure}
    \centering
    \includegraphics[width=\linewidth]{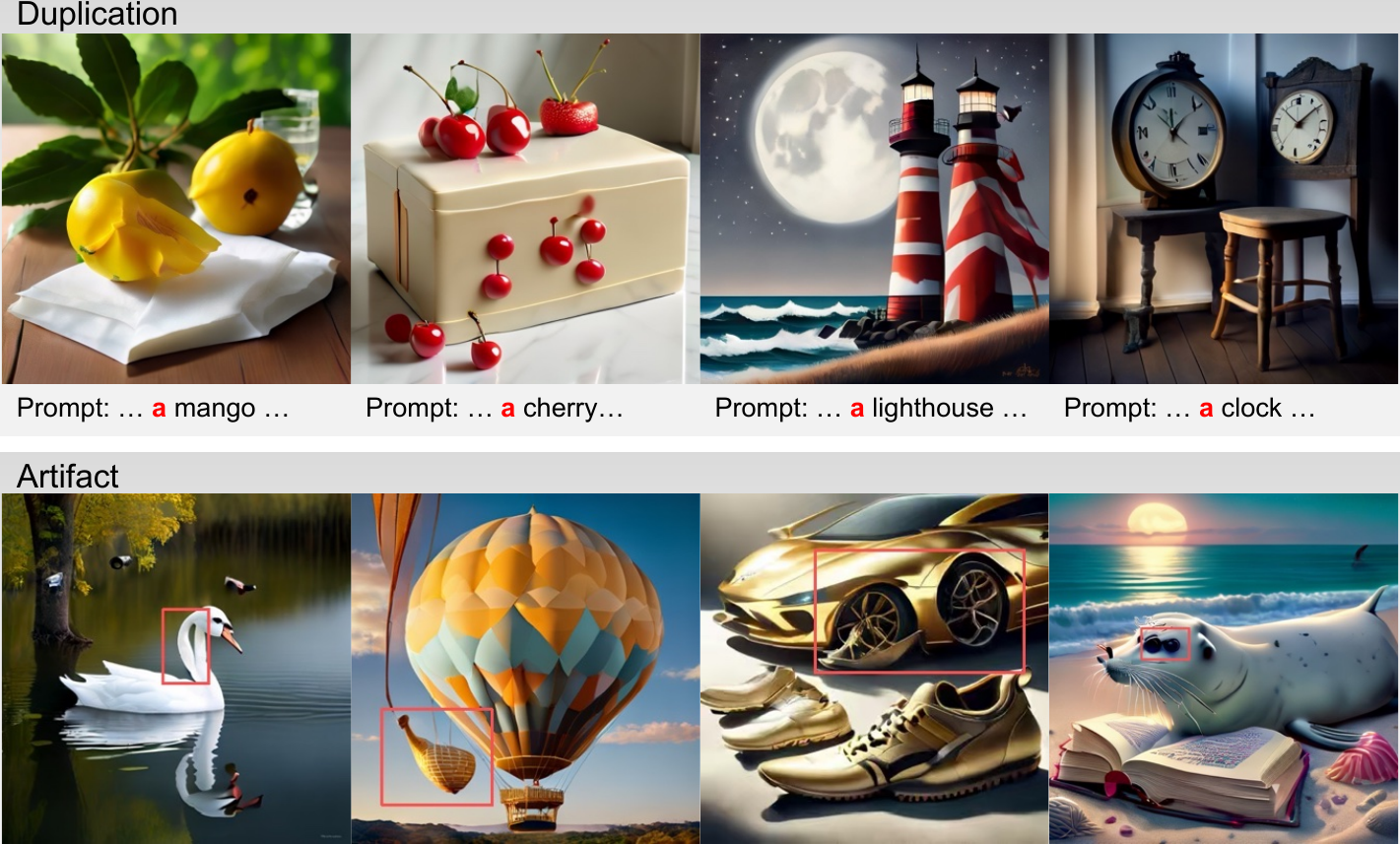}
    \caption{\textbf{Examples of the multi-modality problem in \dimo.} In a single step, \dimo struggles to jointly model the distribution of multiple tokens, causing duplication or even artifacts.}
    \label{fig:dimo}
\end{figure}

It is noteworthy that although the one-step model \dimo achieves extremely fast speed, the image quality is unacceptable in many cases.
This is due to the multi-modality problem~\cite{nat}. The MIGM can only approximate the joint distribution of multiple tokens with the product of the marginal distribution of each token.
This will cause severe performance degradation if the model tries to unmask too many tokens in a single step. \cref{fig:dimo} presents some examples.
In the first row, when prompted to generate a single object, \dimo generates duplication; in the second row, it generates the same parts of an object in different locations, causing artifacts.
Therefore, currently reducing the number of generation steps for MIGMs is particularly difficult, while \method can reduce the number of costly full steps, behaving like pseudo few-step generation.

\paragraph{Human Study.}
To further validate the maintained perceptual quality of DiMOO-Shortcut results, we conduct a human study on the Rapidata platform\footnote{https://www.rapidata.ai/.}, asking humans to compare the quality of images before and after acceleration by DiMOO-Shortcut.
Each trial presents an image generated by vanilla Lumina-DiMOO and the corresponding image generated by DiMOO-Shortcut, with a fixed instruction: \textit{Which image has better overall visual quality?} 
For each pair, three annotators rate independently, and the result is decided by majority voting. All annotators are vetted beforehand and participate on a voluntary basis. 
We report the win rates of DiMOO-Shortcut under different budgets across all the 1065 pairs.
\cref{fig:humanstudy} shows the results. In nearly half the cases, DiMOO-Shortcut with $B=14$ and $4.0\times$ speedup is considered better. Even with $B=9$ and $5.8\times$ speedup, the win rate still approaches 40\%.
These results demonstrate that DiMOO-Shortcut can indeed largely accelerate generation with minimal loss of quality.

% \subsection{Zero-Shot Transfer to Image-to-Image Generation using Lumina-DiMOO}

% \begin{figure}
%     \centering
%     \includegraphics[width=\linewidth]{figs/ca.pdf}
%     \caption{\textbf{Comparison of DiMOO-Shortcut with and without incorporating sampling information.} The right two images are over-smoothed.}
%     \label{fig:ca}
% \end{figure}

% \begin{table*}
% \begin{minipage}{0.72\linewidth}
% \centering
% \captionof{table}{\textbf{Performance of different variants of DiMOO-Shortcut.} $D$ is the number of cross-attention and self-attention layers. $R$ is the bottleneck ratio. ``No cross attention'' means replacing the cross-attention layer with a self-attention layer.\label{tab:ablation}}
% % \resizebox{1.01\linewidth}{!}{
% \input{tabs/ablation}
% % }
% \end{minipage}
% \hfill
% \begin{minipage}{0.25\linewidth}
% \includegraphics[width=\linewidth]{figs/ca.pdf}
% \captionof{figure}{Comparison of DiMOO-Shortcut with and without sampling information.\label{fig:ca}}
% \end{minipage}
% \end{table*}

\subsection{Ablation Study}
\label{sec:ablation}
% We ablate the model design on DiMOO-Shortcut to validate our two core assumptions. The results are shown in \cref{fig:ablation} and Appendix~F. All variants are trained for over 12 hours using 4 H200 GPUs, matching the default setting.
We ablate the model design on DiMOO-Shortcut to validate our two core assumptions. The results are shown in \cref{fig:ablation} and Appendix~\ref{supp:sec:ablation}. All variants are trained for over 12 hours using 4 H200 GPUs, matching the default setting.

\begin{figure}
    \centering
    \includegraphics[width=\linewidth]{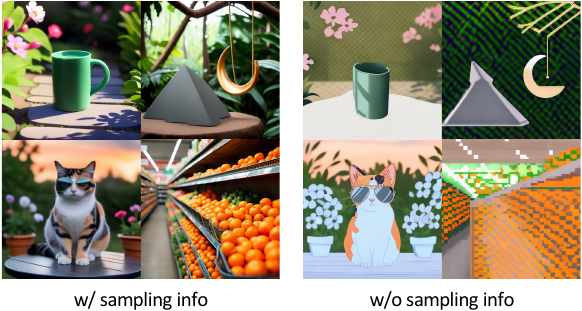}
    \caption{Comparison of DiMOO-Shortcut with and without sampling information.}
    \label{fig:ca}
\end{figure}

\paragraph{Importance of incorporating sampling information.}
The feature dynamics are controlled by the sampling, hence it is a must to incorporate this information to predict features. We implement this via the cross-attention module.
To investigate its influence, we replace the cross-attention with self-attention so that the model cannot attend to sampled tokens.
As \cref{fig:ablation} shows, removing the cross-attention drastically degrades the performance.
\cref{fig:ca} presents some examples. The shortcut model without cross-attention tends to output over-smoothed images, as it is forced to predict the expectation over all possible sampling results.

\paragraph{Model complexity.}
The complexity of the shortcut model should strike a balance. Too low complexity cannot accurately model feature dynamics, like existing caching-based methods, while too high complexity wastes computation, like the base model itself. 
To investigate this, we modify the model size by altering the bottleneck ratio $R$ and the number of cross-attention and self-attention layers $D$ (the setting adopted in the experiments in \cref{sec:maskgit} and \ref{sec:dimoo} is $D=1,R=2$). The variants' performance is shown in \cref{fig:ablation}, with the configurations and parameter numbers indicated in the legend.
Although the heavier variants sometimes surpass the default under the same budget, the performance gain cannot compensate for the increased latency.
On the other hand, making the model lighter compromises performance, which cannot be remedied by increasing the budget.
Therefore, among these variants, our default setting is still Pareto optimal, at a sweet spot of model complexity.

%% file: tabs/maskgit.tex
\begin{tabular}{ll|cc|c}
\toprule
Method & Configuration & Latency (ms)$\downarrow$ & Speedup$\uparrow$ & FID$\downarrow$ \\
\midrule
\multirow{6}{*}{Vanilla} & 8 steps & 26.1 & 1.92$\times$ & 9.91 \\
 & 9 steps & 29.4 & 1.70$\times$ & 8.86 \\
 & 11 steps & 35.9 & 1.40$\times$ & 7.90 \\
 & 13 steps & 42.5 & 1.18$\times$ & 7.64 \\
 & 15 steps & 50.1 & 1.00$\times$ & 7.60 \\
\midrule
\multirow{5}{*}{Shortcut} & 15 steps, $B=7$ & 25.9 & 1.94$\times$ & 8.90 \\
 & 15 steps, $B=8$ & 28.8 & 1.74$\times$ & 8.16 \\
 & 32 steps, $B=8$ & 33.7 & 1.49$\times$ & 7.30 \\
 & 32 steps, $B=9$ & 36.8 & 1.36$\times$ & \underline{6.97} \\
 & 32 steps, $B=12$ & 45.9 & 1.09$\times$ & \textbf{6.84} \\
\bottomrule
\end{tabular}

%% file: tabs/main.tex
\begin{tabular}{ll|ccc|ccc}
\toprule
Method & Configuration & Latency (s)$\downarrow$ & Speedup$\uparrow$ & Memo. (GB)$\downarrow$ & ImageReward$\uparrow$ & CLIPScore$\uparrow$ & UniPercept-IQA$\uparrow$ \\
\midrule
Vanilla & 64 steps & 23.10 & 1.00$\times$ & 19.07 & 0.91 {\scriptsize (+0.00)} & 34.46 {\scriptsize (+0.00)} & 71.07 {\scriptsize (+0.00)} \\
\midrule
 \multirow{4}{*}{\makecell[l]{Few-step \\ Vanilla}} & 32 steps & 11.60 & 1.99$\times$ & \multirow{4}{*}{19.07} & 0.89 {\scriptsize (-0.02)} & 34.35 {\scriptsize (-0.11)} & \underline{71.00 {\scriptsize (-0.07)}} \\
 & 16 steps & 5.77 & 4.00$\times$ & & 0.75 {\scriptsize (-0.16)} & 34.19 {\scriptsize (-0.27)} & 68.61 {\scriptsize (-2.46)} \\
 & 14 steps & 5.04 & 4.58$\times$ & & 0.70 {\scriptsize (-0.21)} & 34.10 {\scriptsize (-0.36)} & 67.84 {\scriptsize (-3.23)} \\
 & 13 steps & 4.68 & 4.94$\times$ & & 0.67 {\scriptsize (-0.24)} & 34.04 {\scriptsize (-0.42)} & 67.50 {\scriptsize (-3.57)} \\
\midrule
\multirow{2}{*}{ML-Cache} & $(w, r, c) = (0.1, 5, 0.9)$ & 10.40 & 2.22$\times$ & \multirow{2}{*}{24.25} & \textbf{0.90 {\scriptsize (-0.01)}} & 34.47 {\scriptsize (+0.01)} & 70.68 {\scriptsize (-0.39)} \\
 & $(w, r, c) = (0.0, 5, 0.9)$ & 9.10 & 2.54$\times$ & & 0.66 {\scriptsize (-0.25)} & 34.04 {\scriptsize (-0.42)} & 66.81 {\scriptsize (-4.26)} \\
\midrule
\multirow{2}{*}{ReCAP*} & $(u, T, T') = (0, 8, 40)$ & 7.05 & 3.28$\times$  & \multirow{2}{*}{23.19} & 0.80 {\scriptsize (-0.11)} & \textbf{34.54 {\scriptsize (+0.08)}} & 69.09 {\scriptsize (-1.98)} \\
 & $(u, T, T') = (0, 8, 24)$ & 5.42 & 4.26$\times$  & & 0.77 {\scriptsize (-0.14)} & 34.36 {\scriptsize (-0.10)} & 68.63 {\scriptsize (-2.44)} \\
\midrule
\multirow{2}{*}{dLLM-Cache*} & $(K_p, K_r) = (10, 5)$ & 15.67 & 1.47$\times$ & \multirow{2}{*}{35.58} & 0.79 {\scriptsize (-0.12)} & \underline{34.48 {\scriptsize (+0.02)}} & 70.76 {\scriptsize (-0.31)} \\
 & $(K_p, K_r) = (64, 16)$ & 14.30 & 1.62$\times$ & & 0.43 {\scriptsize (-0.48)} & 33.78 {\scriptsize (-0.68)} & 65.98 {\scriptsize (-5.09)} \\
\midrule
\multirow{2}{*}{TaylorSeer*} & $(\mathcal{N},O)=(4,2)$ & 8.71 & 2.65$\times$ & \multirow{2}{*}{31.56} & 0.81 {\scriptsize (-0.10)} & 34.41 {\scriptsize (-0.05)} & 69.87 {\scriptsize (-1.20)} \\
 & $(\mathcal{N},O)=(8,2)$ & 5.99 & 3.86$\times$ & & 0.37 {\scriptsize (-0.54)} & 33.62 {\scriptsize (-0.84)} & 61.28 {\scriptsize (-9.79)} \\
\midrule
\multirow{1}{*}{\dimo} &  & 0.07 & 330.00$\times$ & 6.28 & 0.43 {\scriptsize (-0.48)} & 33.99 {\scriptsize (-0.47)} & 66.40 {\scriptsize (-4.67)} \\
\midrule
\multirow{3}{*}{Shortcut} & $(N,B)=(64,14)$ & 5.76 & 4.01$\times$ &  \multirow{3}{*}{20.05} & \textbf{0.90 {\scriptsize (-0.01)}} & \underline{34.48 {\scriptsize (+0.02)}} & \textbf{71.25 {\scriptsize (+0.18)}} \\
 & $(N,B)=(64,11)$ & 4.70 & 4.91$\times$ &  & 0.87 {\scriptsize (-0.04)} & 34.39 {\scriptsize (-0.07)} & 70.80 {\scriptsize (-0.27)} \\
 & $(N,B)=(64,9)$ & 3.99 & 5.79$\times$ &  & 0.83 {\scriptsize (-0.08)} & 34.37 {\scriptsize (-0.09)} & 70.68 {\scriptsize (-0.39)} \\
\bottomrule
\end{tabular}

%% file: secs/5-conclusion.tex
\section{Conclusion}
This paper presents \method, which achieves remarkable acceleration with minimal performance drop.
We identify that the sampling process of MIGMs wastes much information in the continuous features, hence leveraging this information may reduce the computational complexity.
For existing related methods that use previous features to approximate future ones, we point out that they suffer from limited expressivity and the fatal neglect of sampling information.
Accordingly, we propose to learn a lightweight model that incorporates both previous features and newly sampled tokens to learn the dynamics of latent feature evolution.
The learned model takes a shortcut in the latent space that skips the heavy base model to improve efficiency.
Experiments on MaskGIT and Lumina-DiMOO demonstrate the effectiveness and rationale of \method.
We hope this paper could provide more insights into the computation paradigm of MIGMs, especially about its inherent redundancy.

%% file: supp/feature_layer.tex
\section{Selecting which Layer's Feature to Model}
\label{supp:sec:feature-layer}

% \begin{figure*}[b]
%     \centering
% \begin{minipage}{0.25\linewidth}
% \includegraphics[width=\linewidth]{figs/layer_selection.png}
% \captionof{figure}{\textbf{Feature cosine similarity between consecutive steps vs. layer depth.} The line represents the average over all tokens, and the shade represents the range from 25\% to 75\% in all tokens.\label{supp:fig:layer_sel}}

% \end{minipage}
% \hfill
% \begin{minipage}{0.6\linewidth}
%     \includegraphics[width=\linewidth]{figs/attempts.png}
%     \captionof{figure}{\textbf{Comparison of different training strategies.} ``Dist matching'' means adding a KL divergence term to the loss. ``Rollout2/3'' means rolling out 2/3 steps using the shortcut model in training.\label{supp:fig:attempts}}
% \end{minipage}
% \end{figure*}

\begin{figure}[b]
    \centering
    \includegraphics[width=0.7\linewidth]{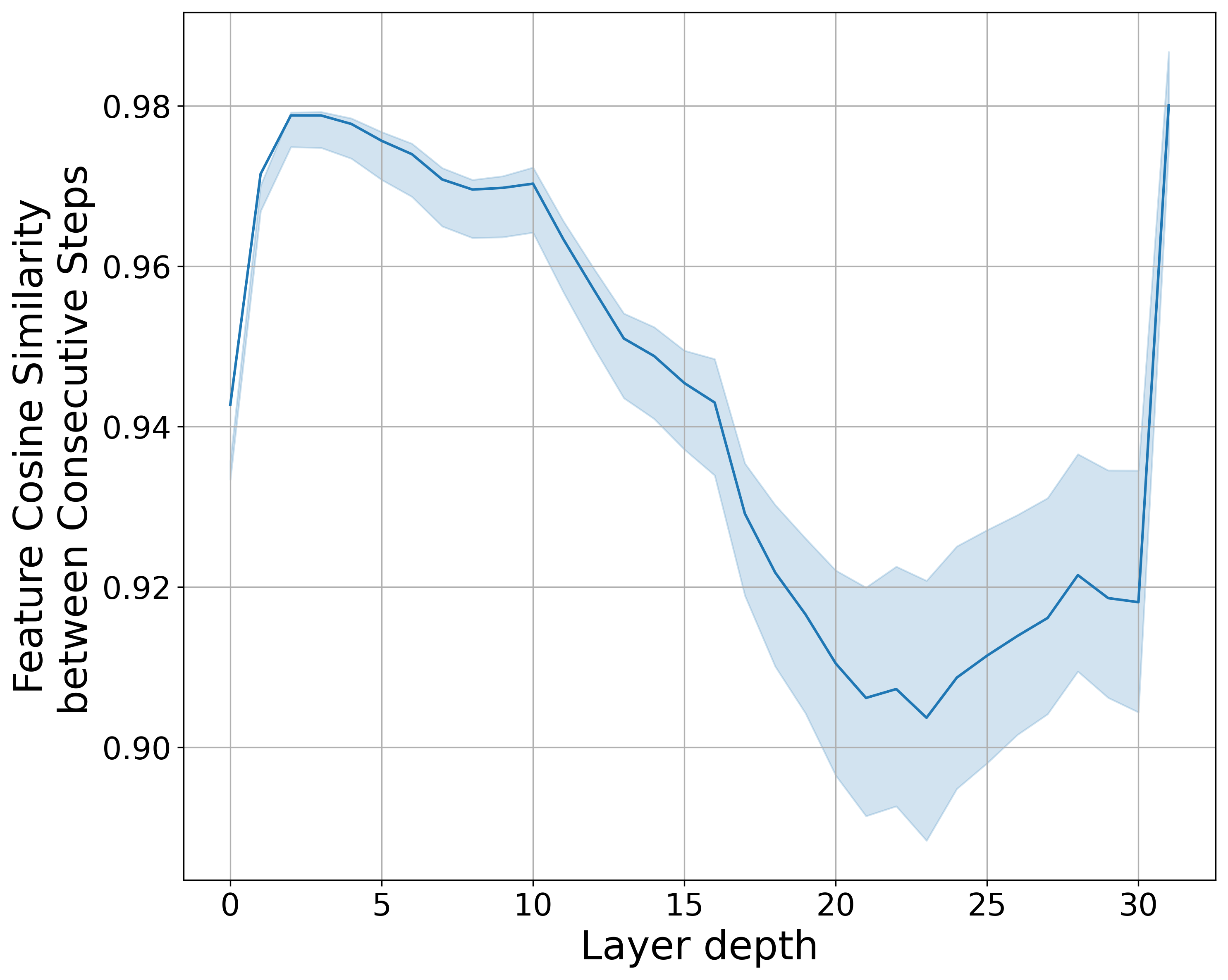}
    \caption{\textbf{Feature cosine similarity between consecutive steps vs. layer depth.} The line represents the average over all tokens, and the shade represents the range from 25\% to 75\% in all tokens.}
    \label{supp:fig:layer_sel}
\end{figure}

We inspect which layer's feature is the most suitable for modeling by our \method.
In order for \method to achieve higher acceleration rates, we hope to use the last-layer feature.
In order for the dynamics to be amenable to learning, we hope the cosine similarity between consecutive steps is the highest.
\cref{supp:fig:layer_sel} plots the similarity vs. the layer depth for an exemplary sample generated by Lumina-DiMOO.
It turns out that the two conditions are achieved simultaneously: the last-layer feature exhibit the most stable dynamics.
Therefore, we choose this feature as the modelling target of \method.
A possible explanation is that at the end of the neural network, as the feature directly dominates the generated distribution, it must be close to the training target, hence converging to a similar space across different steps.

%% file: supp/attempts.tex
\section{More Exploration of Training Strategies}
\label{supp:sec:attempts}
Beyond a simple MSE loss, we have also tried other strategies for training DiMOO-Shortcut.
% 
% \begin{itemize}
\paragraph{Adding distribution matching term to loss.} We add the KL divergence between the logits from the shortcut model's features and the base model's features:
    $$D_{\text{KL}}(\text{softmax}(H(\hat{\boldsymbol{f}}))\ \Vert \ \text{softmax}(H(\boldsymbol{f}))),$$
    where $H$ is the frozen classification head of the base model.
\paragraph{Exposing shortcut model to its prediction.} At inference time, the input to the shortcut model may be its own feature prediction instead of the true features from the base model. This imposes a risk of exposure bias. Therefore, we propose that in training, let the shortcut model roll out for several steps, and then feed the resulting features into the model.
% \end{itemize}
\par
\begin{figure}[b]
    \centering
    \includegraphics[width=\linewidth]{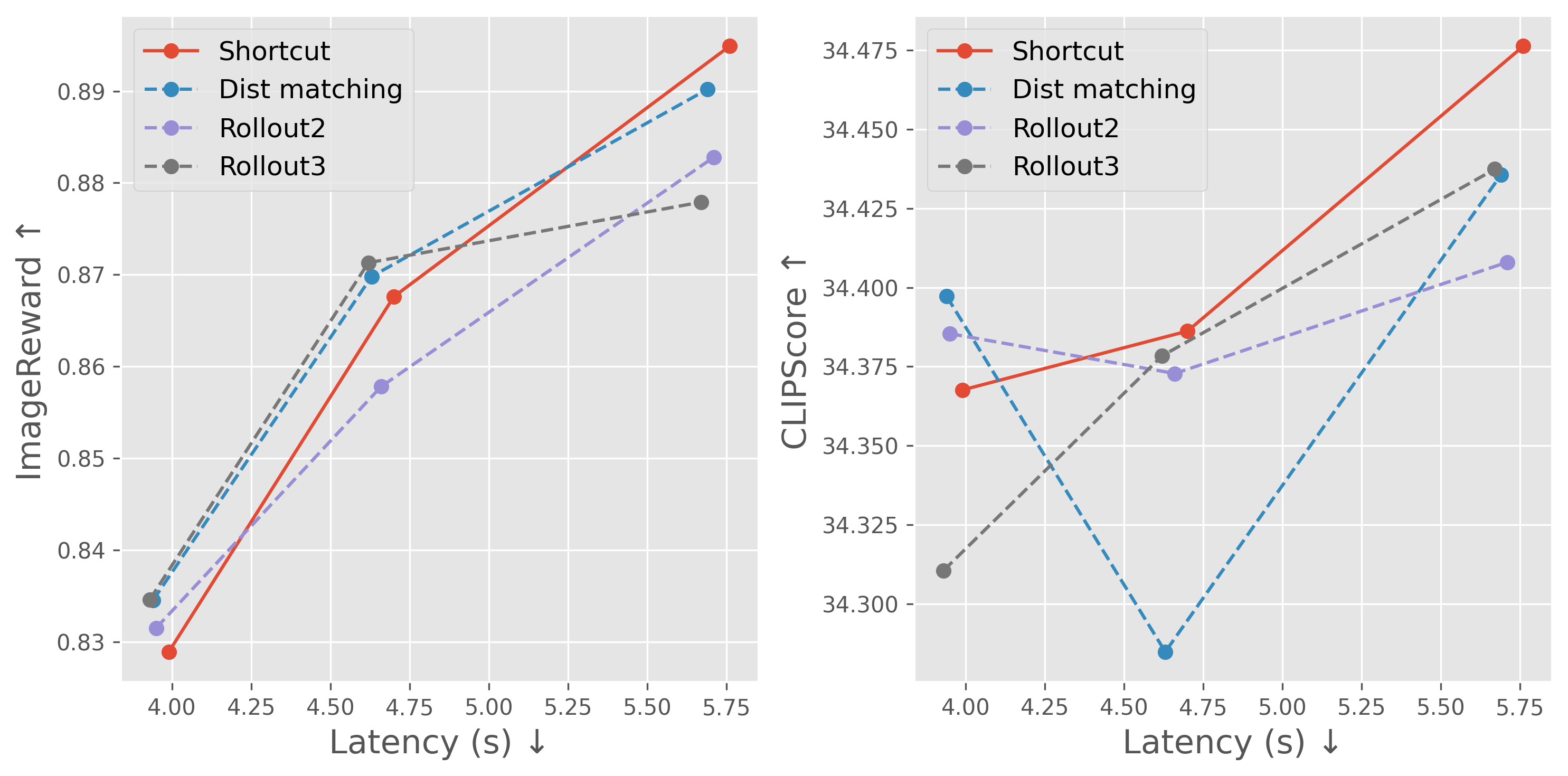}
    \caption{\textbf{Comparison of different training strategies on DiMOO-Shortcut.} ``Shortcut'' means the standard setting. ``Dist matching'' means adding a KL divergence term to the loss. ``Rollout2/3'' means rolling out 2/3 steps using the shortcut model in training.}
    \label{supp:fig:attempts}
\end{figure}

\medskip
Although these methods seem to make sense, they do not bring notable benefits, as shown in \cref{supp:fig:attempts}.
The reason may be that the dynamics are smooth enough and thus the target is easy enough to be learned by an MSE loss, which is in line with our core assumption.

\begin{table*}[h]
    \centering
    \caption{\textbf{Performance of different variants of DiMOO-Shortcut.} $D$ is the number of cross-attention and self-attention layers. $R$ is the bottleneck ratio. ``No cross attention'' means replacing the cross-attention layer with a self-attention layer.}
    \vspace{-5pt}
    \label{supp:tab:ablation}
    \input{tabs/ablation}
\end{table*}

%% file: tabs/ablation.tex
\begin{tabular}{lc|c|ccc}
\toprule
Method & Budget & Latency (s)$\downarrow$ & ImageReward$\uparrow$ & CLIPScore$\uparrow$ & UniPercept-IQA$\uparrow$ \\
\midrule
\multirow{3}{*}{\makecell[l]{$D=1,R=2$\\ (default)}} & $14$ & 5.76 & 0.90 & 34.48 & 71.25 \\
 & $11$ & 4.70 & 0.87 & 34.39 & 70.80 \\
 & $9$ & 3.99 & 0.83 & 34.37 & 70.68 \\
\midrule
\multirow{3}{*}{$D=2,R=2$} & $14$ & 6.20 & 0.90 & 34.48 & 70.75 \\
 & $11$ & 5.17 & 0.86 & 34.34 & 70.62 \\
 & $9$ & 4.49 & 0.85 & 34.31 & 70.40 \\
\midrule
\multirow{3}{*}{$D=1,R=1$} & $14$ & 6.24 & 0.88 & 34.38 & 70.65 \\
 & $11$ & 5.20 & 0.85 & 34.30 & 70.51 \\
 & $9$ & 4.50 & 0.82 & 34.25 & 70.39 \\
\midrule
\multirow{4}{*}{$D=1,R=4$} & $16$ & 6.19 & 0.87 & 34.34 & 70.85 \\
 & $14$ & 5.56 & 0.85 & 34.39 & 70.87 \\
 & $11$ & 4.40 & 0.83 & 34.24 & 70.72 \\
 & $9$ & 3.70 & 0.79 & 34.23 & 69.57 \\
\midrule
\multirow{3}{*}{\makecell[l]{$D=1,R=2$,\\ no cross attention}} & $14$ & 5.84 & 0.69 & 33.89 & 68.09 \\
 & $11$ & 4.78 & 0.58 & 33.67 & 65.86 \\
 & $9$ & 4.06 & 0.46 & 33.34 & 63.85 \\
\bottomrule
\end{tabular}

%% file: supp/more_eval.tex
\section{More Evaluation}
\label{supp:sec:eval}
\paragraph{Composition.} \cref{supp:tab:eval} lists performance of DiMOO-Shortcut on two popular text-to-image benchmarks, GenEval~\cite{geneval} and DPG-Bench~\cite{dpg}, which focuses on composition. Each evaluation is run 4 times as officially recommended. The scores are maintained on average but fluctuate widely, providing limited reference value. Besides, we observe that these scores do not align well with perceptual quality. So in the main text we report the continuous scores that are better suited to measuring nuance of generation quality, following common practice.

\begin{figure}[b]
    \centering
    \includegraphics[width=\linewidth]{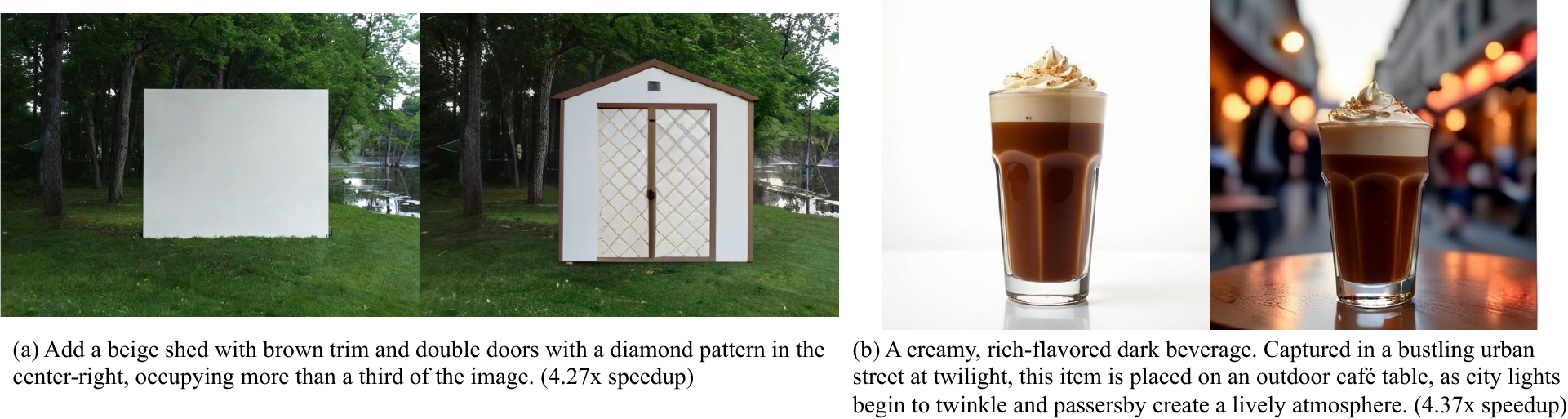}
    % \vspace{-5pt}
    \caption{Editing examples (no cherry-picking).}
    \label{supp:fig:edit}
\end{figure}

\paragraph{Generalization.} The evaluation above of DiMOO-Shortcut is under the recommended inference setting and a specific prompt distribution (from DPG-Bench). 
To evaluate generalization across prompt distribution, mask schedule, CFG scale, and resolution, we conduct controlled experiments and report the results in \cref{supp:tab:gene}. Only the setting in the table head differs from the default. Besides, for 512 resolution, \#steps is 16 and the budget is 8 with $2.0\times$ speedup; for other settings, the budget is 14. It turns out that our method is robust to different settings. 
\cref{supp:fig:edit} also shows examples of unseen image-to-image editing tasks, suggesting that the learned latent dynamics can generalize for different image generation tasks.

\begin{table}[tb]
    \centering
    \caption{Results on GenEval and DPG-Bench.}
    \vspace{-5pt}
    % \resizebox{0.95\linewidth}{!}{
    \begin{tabular}{l|c|cc}
    \toprule
      Method   & Speedup$\uparrow$ & GenEval$\uparrow$ & DPG-Bench$\uparrow$ \\
      \midrule
      Vanilla (64 steps) & $1.00\times$  &  $86.4 \pm 1.9$ & $85.6 \pm 0.1$  \\
      Vanilla (16 steps) & $4.00\times$ & $85.6 \pm 0.8$  & $81.6\pm 0.8$  \\
      Shortcut ($B=14$)  & $4.01\times$ & $86.5\pm 0.8$  & $85.2 \pm 0.2$   \\
      Shortcut ($B=11$)  & $4.91\times$ & $86.1 \pm 1.6$  & $85.0 \pm 0.4$  \\
      Shortcut ($B=9$)  & $5.79\times$ & $83.6 \pm 1.1$  & $84.6 \pm 0.4$  \\
      \bottomrule
    \end{tabular}
    % }
    % \vspace{-10pt}
    \label{supp:tab:eval}
\end{table}

\begin{table}[tb]
    \centering
    % \vspace{-10pt}
    \caption{Performance before/after acceleration under different settings.}
    \vspace{-5pt}
    \resizebox{\linewidth}{!}{
    \begin{tabular}{l|cccc}
    \toprule
      Metric & Prompt: GenEval & Resolution: 512 & CFG: 2 & Mask schedule: arccos  \\
      \midrule
      CLIPScore  & 33.44/33.42 & 34.44/34.33 & 33.90/33.89 & 34.25/34.06 \\
      UniPercept  & 72.22/72.37 & 66.54/66.49 & 68.50/68.91 &  68.86/67.01 \\
      \bottomrule
    \end{tabular}
    }
    \label{supp:tab:gene}
\end{table}

%% file: supp/more_visualization.tex
\section{More Visualization}
\label{supp:sec:more_visualization}
Fig.~\ref{supp:fig:pic2}-\ref{supp:fig:pic9} present more examples of images generated by DiMOO-Shortcut and other methods, demonstrating that our method could consistently generate high-quality images with high speed.

%% file: supp/fail.tex
\section{Failure Cases}
\begin{figure}[tb]
    \centering
    \includegraphics[width=\linewidth]{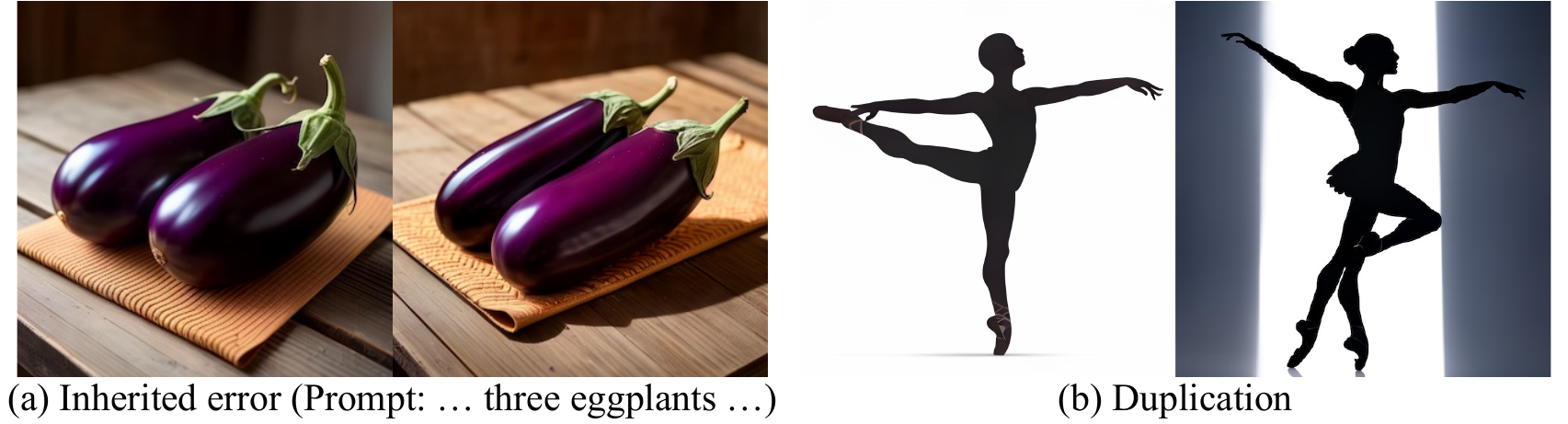}
    \vspace{-10pt}
    \caption{Failure cases. Left: vanilla; Right: shortcut.}
    \label{supp:fig:fail}
\end{figure}

\cref{supp:fig:fail} presents failure cases. In (a), the shortcut model inherits the base model's error, as the base model imposes an upper bound. In (b), the shortcut generates duplication, indicating that the multimodality problem has not yet been fully resolved.

%% file: supp/ablation_tab.tex
\section{Complete Results of Ablation Study}
\label{supp:sec:ablation}
\cref{supp:tab:ablation} lists the metric values of different variants of DiMOO-Shortcut. Among them, the default setting $N = 1, R = 2$ with cross-attention achieves the optimal quality-speed tradeoff.

%% file: supp/theoretical_support.tex
\section{Theoretical Support}
Denote the ground truth of shortcut target $f_{i+1}-f_{i}$ by $R(f_{i},x_{i})$\footnote{For brevity we write time $t_k$ as $k$ henceforth and omit time in $S_\theta$.}. Assume when $\Vert \hat{f}_i - f_i \Vert \leq \delta_L,\Vert S_\theta(\hat{f}_i,x_i)-S_\theta(f_i,x_i)\Vert \leq L\Vert \hat{f}_i-f_i\Vert$ for some $L>0$ (\textit{local Lipschitz regularity assumption}). Assume $\Vert S_\theta(f_i,x_{i})-R(f_i,x_{i}) \Vert \leq \epsilon$ (\textit{bounded local approximation error assumption}). Also assume when $\Vert\hat{f}_{i} -f_{i}\Vert \leq \delta_S$, the $\hat{x}_{i}$ sampled from $K(\cdot | x_{i-1},\text{softmax}(H(\hat{f}_{i})),N_{i})$ can be seen as a high-probability draw from $K(\cdot | x_{i-1},\text{softmax}(H(f_{i})),N_{i})$ so that $\hat{x}_{i}$ can be treated as $x_i$ (\textit{sampling admissibility assumption}). 
Consecutively applying $M$ shortcut steps from $f_j$. If $\Vert\hat{f}_{j+m} -f_{j+m}\Vert \leq \min(\delta_L,\delta_S)\eqqcolon\delta$ for $m\in [M-1]$, then $\hat{x}_{j+m},m\in[M-1]$ is a plausible trajectory and therefore
\[
\begin{aligned}
&\Vert\hat{f}_{j+M} -f_{j+M}\Vert\\
=&\ \Vert\hat{f}_{j+M-1} - f_{j+M-1} + S_\theta(\hat{f}_{j+M-1},x_{j+M-1})-R(f_{j+M-1},x_{j+M-1})\Vert \\
\leq&\ \Vert\hat{f}_{j+M-1} - f_{j+M-1}\Vert \\
& \quad + \Vert S_\theta(\hat{f}_{j+M-1},x_{j+M-1})-S_\theta(f_{j+M-1},x_{j+M-1})\Vert \\
& \quad + \Vert S_\theta(f_{j+M-1},x_{j+M-1})-R(f_{j+M-1},x_{j+M-1})\Vert \\
\leq&\ \epsilon + (1+L)\Vert\hat{f}_{j+M-1} -f_{j+M-1}\Vert.
\end{aligned}
\]
% $\Vert\hat{f}_{i+M} -f_{i+M}\Vert \leq \epsilon + 2\Vert\hat{f}_{i+M-1} -f_{i+M-1}\Vert$. 
It follows that 
$$\Vert\hat{f}_{j+M} -f_{j+M}\Vert \leq \epsilon\sum_{n=0}^{M-1}(1+L)^n = \epsilon\frac{(1+L)^{M}-1}{L}.$$
So by induction, if $\epsilon\frac{(1+L)^{T-1}-1}{L} \leq \delta$, then refreshing every $T$ steps suppresses the accumulation error and keeps the sample in a local plausible regime as long as the above three assumptions hold.
Note that the tightest local constants above may vary with time, and future work will consider more sophisticated adaptive refresh schedules to improve performance.

At the core of the assumptions is that the map $(f_i,x_i) \mapsto f_{i+1}-f_i$ is less computationally complex than the original map $x_i \mapsto f_{i+1}$. Although it is hard to derive a closed-form expression, we can analyse factors that make the required condition relatively mild.
Let the embedding of $x_i$ be $e_i$, $f_{i+1} = F(e_i)$, and $\Delta_{i}=e_{i}-e_{i-1}$. If $F$'s Jacobian $J_F$ is $\beta$-Lipschitz, then $f_{i+1}-f_i=F(e_{i})-F(e_{i-1})=J_F(e_{i-1})\Delta_i+r_i,\Vert r_i\Vert^2 \leq \frac{\beta}2\Vert\Delta_i\Vert^2$. $\Delta_i$ is nonzero only in the positions of new tokens at step $i$, and the residual term $r_i$ is bounded. That is, this objective is driven by sparse factors. Therefore, conditioned on the rich context $f_i$ and the necessary info $x_i$, predicting $f_{i+1}-f_i$ is likely to have lower complexity compared to recomputation from all tokens. 
These theoretical analyses indicate our method is feasible under mild assumptions. Though the assumptions depend on concrete architectures, tokenizers, and sampling strategies, the experiments have empirically validated the practical effectiveness in various settings.

%% file: supp/more_related_work.tex
\input{supp/exs}
\section{Broader Related Work}
\label{supp:sec:related_work}
There exist broad relations to other topics, including knowledge distillation~\cite{hinton2015distilling,fitnet,selfdistill}, latent prediction~\cite{jepa,ijepa,repa}, and consistency-style acceleration~\cite{consistencymodels,ctm,continuousshortcut}. However, our method differs from them in a non-trivial sense. Rather than directly applying existing methods to MIGMs, we derive a method tailored to specific problems in this area, and the method can be supported by other methodologies in different aspects.

\paragraph{Knowledge distillation and student models.} Although the learning objective is similar in form, the modeled objects differ. In the standard setting, knowledge distillation transfers the teacher's knowledge to the student, with the two models sharing the same input and output, so the knowledge is homogeneous. However, for our shortcut model and the base model, the output is the same but the input is different, so the mapping is distinct, and the learned knowledge is different. In knowledge distillation, the student follows the same way as the teacher; but in our paper, the shortcut model leads to the same objective as the base model in a different way.

\paragraph{Latent prediction.} Some works assume there exists a well-behaved latent space underlying observations, and seek to model it. Our idea includes but is not limited to this perspective. The insight is not only on what to predict (output), but also on where the prediction is from (input). It is the shortcut connecting the two ends that forms our holistic contribution.

\paragraph{Consistency-style acceleration.} The common point is using intermediate features as the supervision signal to learn to skip some computation. The difference is that consistency-style models skip steps and refactor the trajectory, accelerating by reducing steps; in contrast, our method skips the base model and still tracks the original trajectory, accelerating by reducing per-step cost.

%% file: supp/exs.tex
% \begin{figure*}
%     \centering
%     \includegraphics[width=0.91\linewidth]{figs/sup/pic1.pdf}
%     \caption{Prompt: Two multicolored butterflies with delicate, veined wings gently balance atop a vibrant, orange tangerine in a bustling garden. The tangerine, with its glossy, dimpled texture, is situated on a wooden table, contrasting with the greenery of the surrounding foliage and flowers. The butterflies, appearing nearly small in comparison, add a touch of grace to the scene, complementing the natural colors of the verdant backdrop.}
%     \label{supp:fig:pic1}
% \end{figure*}

\begin{figure*}[!b]
    \centering
    \includegraphics[width=0.91\linewidth]{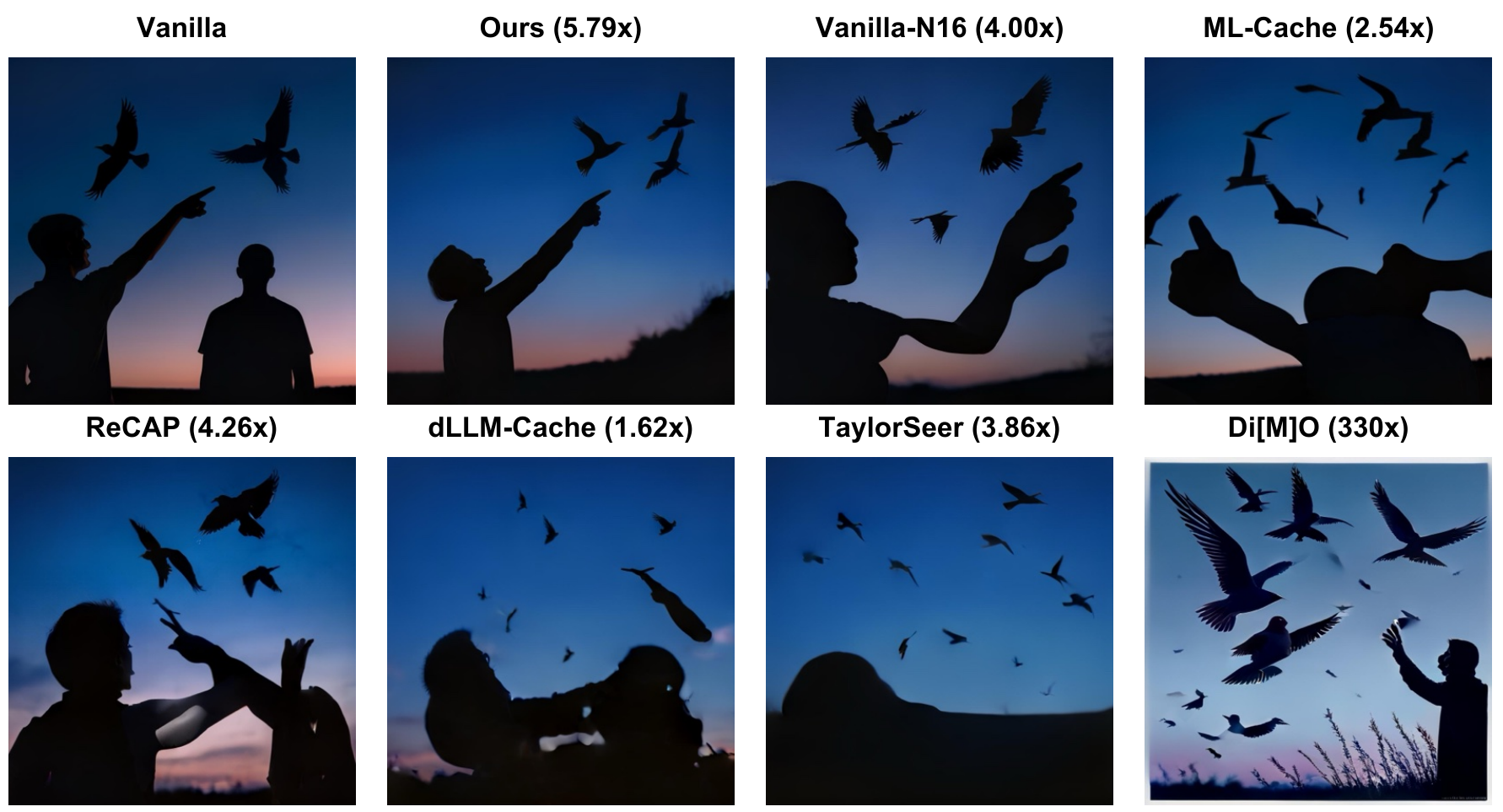}
    \caption{Prompt: During the twilight hour, an individual can be seen extending an arm towards the sky, pointing at a trio of wild birds gliding through the rich deep blue of the early evening sky. The birds' silhouettes contrast distinctly against the fading light, their wings spread wide as they soar. The person is silhouetted against the dusky sky, creating a peaceful scene of human connection with nature.}
    \label{supp:fig:pic2}
\end{figure*}

\begin{figure*}
    \centering
    \includegraphics[width=0.91\linewidth]{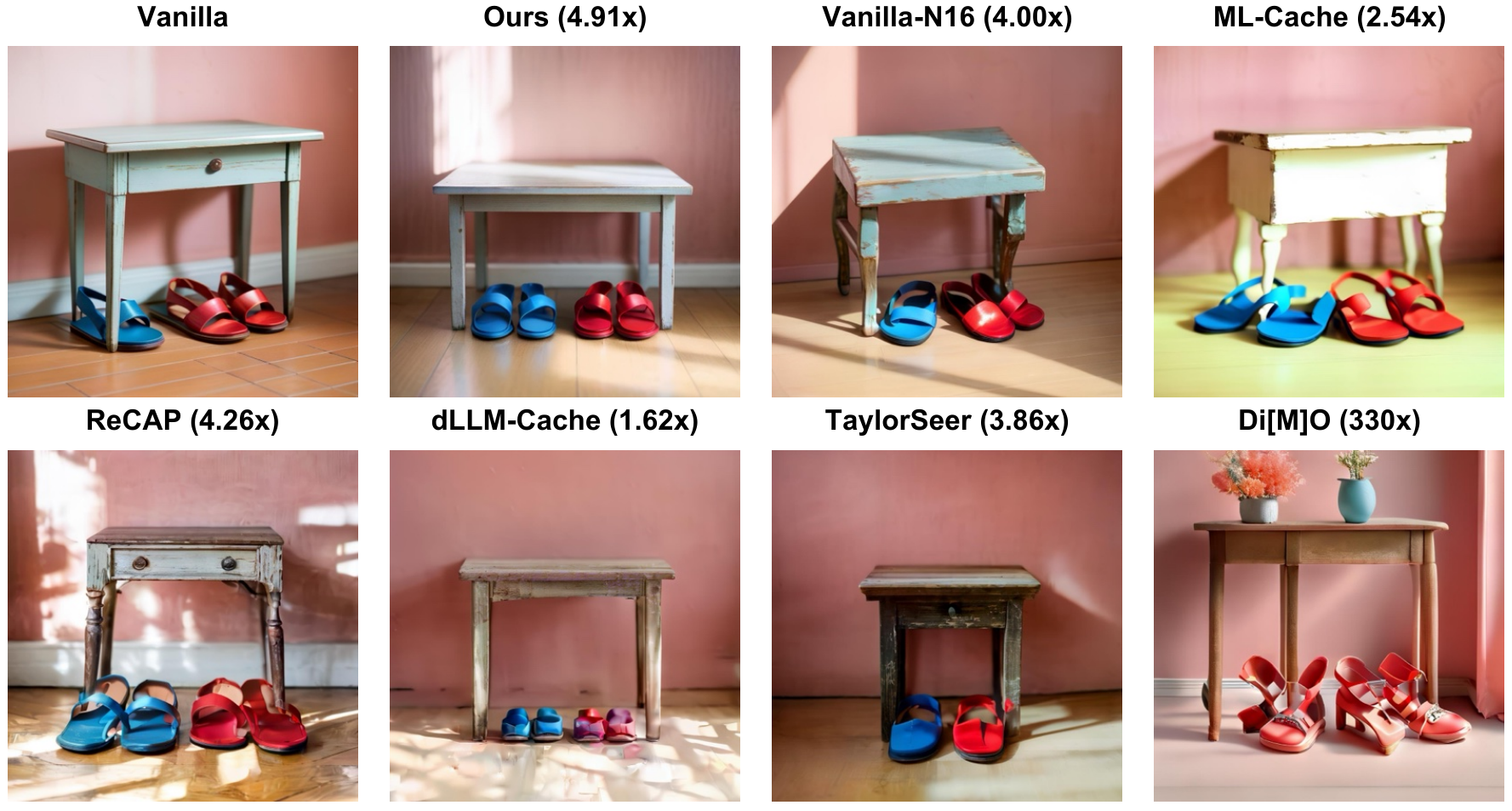}
    \caption{Prompt: In an elegantly simple room, two pairs of sandals—one blue and one red—sit tidily beneath a quaint, square wooden side table featuring a weathered finish that suggests a touch of rustic charm. The side table, which casts a soft shadow onto the textured salmon pink wall behind it, provides a harmonious balance to the vibrant footwear. The floor beneath the table is a polished light hardwood, reflecting a faint glow from the natural light entering the room.}
    \label{supp:fig:pic3}
\end{figure*}

\begin{figure*}
    \centering
    \includegraphics[width=0.91\linewidth]{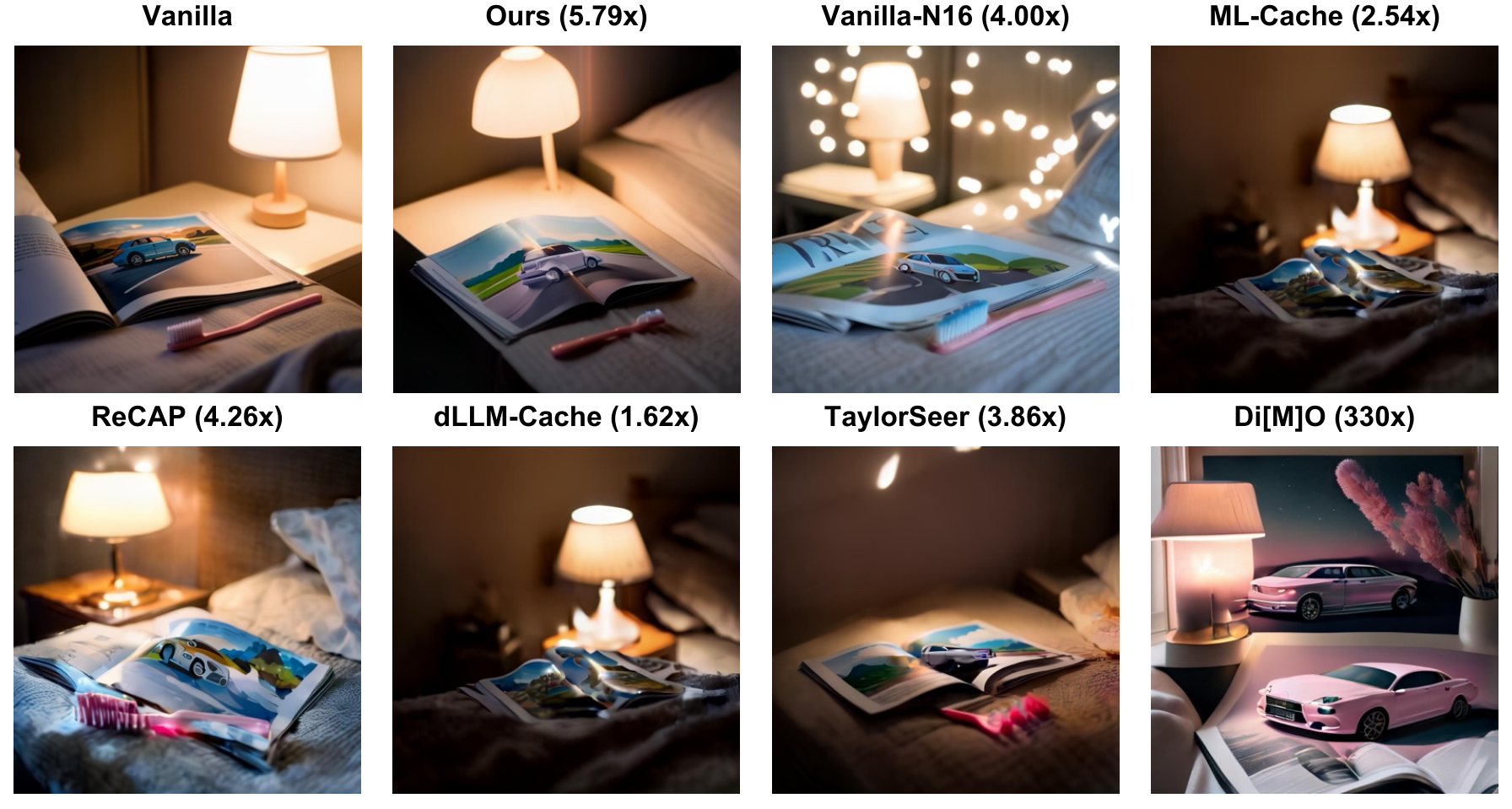}
    \caption{Prompt: Inside a dimly lit room, the low luminance emanates from a bedside lamp casting a soft glow upon the nightstand. There lies a travel magazine, its pages open to a vivid illustration of a car driving along a picturesque landscape. Positioned on the image is a light pink toothbrush, its bristles glistening in the ambient light. Beside the magazine, the textured fabric of the bedspread is just discernible, contributing to the composed and quiet scene.}
    \label{supp:fig:pic4}
\end{figure*}

% \begin{figure*}
%     \centering
%     \includegraphics[width=0.91\linewidth]{figs/sup/pic5.pdf}
%     \caption{Prompt: A delicate silver ring with a sleek band sits gracefully next to a large, dark leather wallet that seems to overflow with cards and receipts. Both items rest on a polished wooden bedside table, whose grain texture subtly complements the metallic luster of the ring. The wallet’s bulky appearance emphasizes the fine simplicity of the ring, highlighting the stark difference in their sizes and designs.}
%     \label{supp:fig:pic5}
% \end{figure*}

\begin{figure*}
    \centering
    \includegraphics[width=0.91\linewidth]{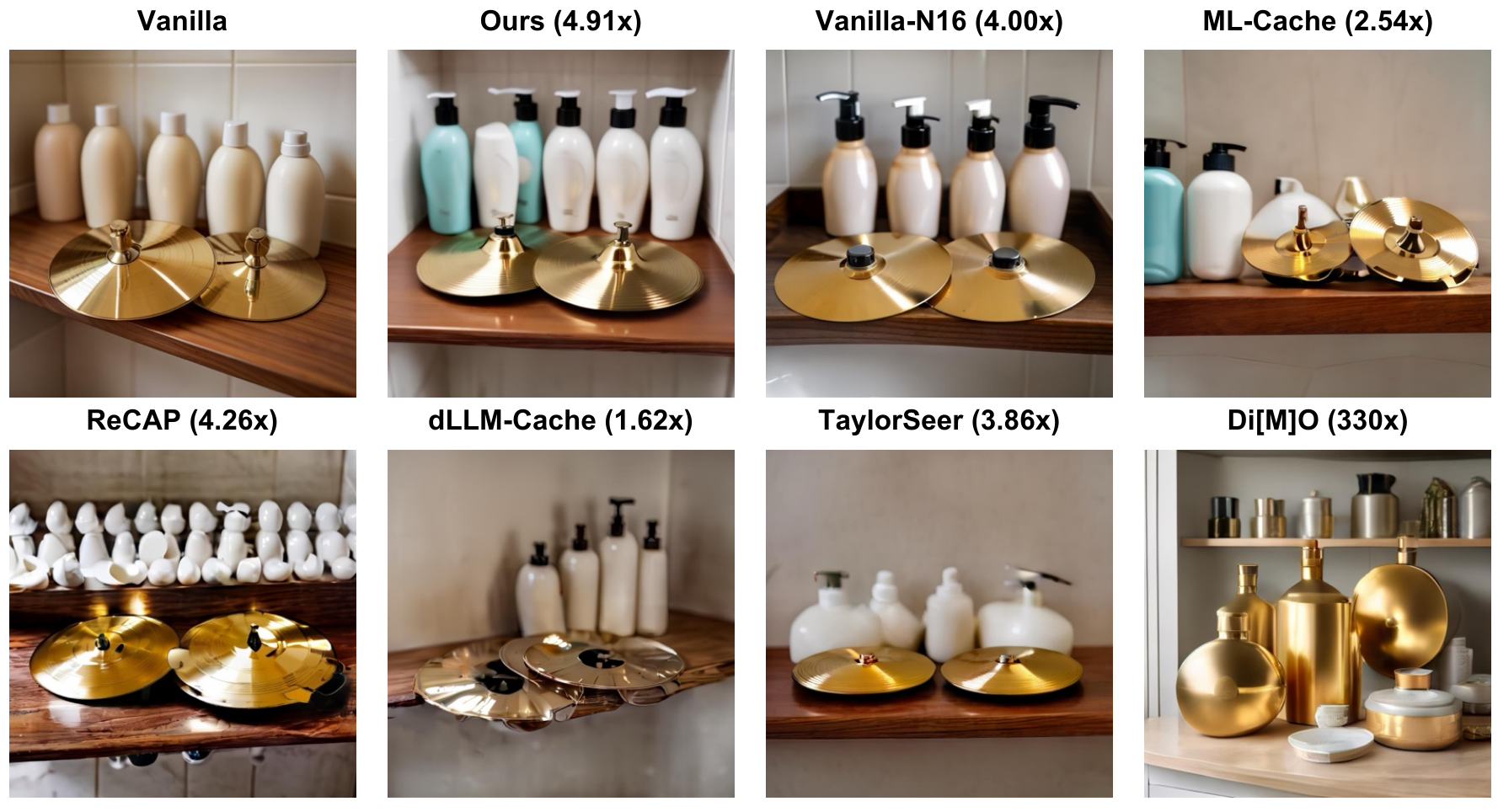}
    \caption{Prompt: A gleaming pair of golden cymbals lies in stark contrast to a collection of round, static bottles of toiletries arranged neatly on a shelf. The metallic sheen of the cymbals is emphasized by the surrounding muted tones of the shampoo and lotion containers. The bathroom shelf upon which they rest is made of polished oak, adding a warm touch to the setting.}
    \label{supp:fig:pic6}
\end{figure*}

\begin{figure*}
    \centering
    \includegraphics[width=0.91\linewidth]{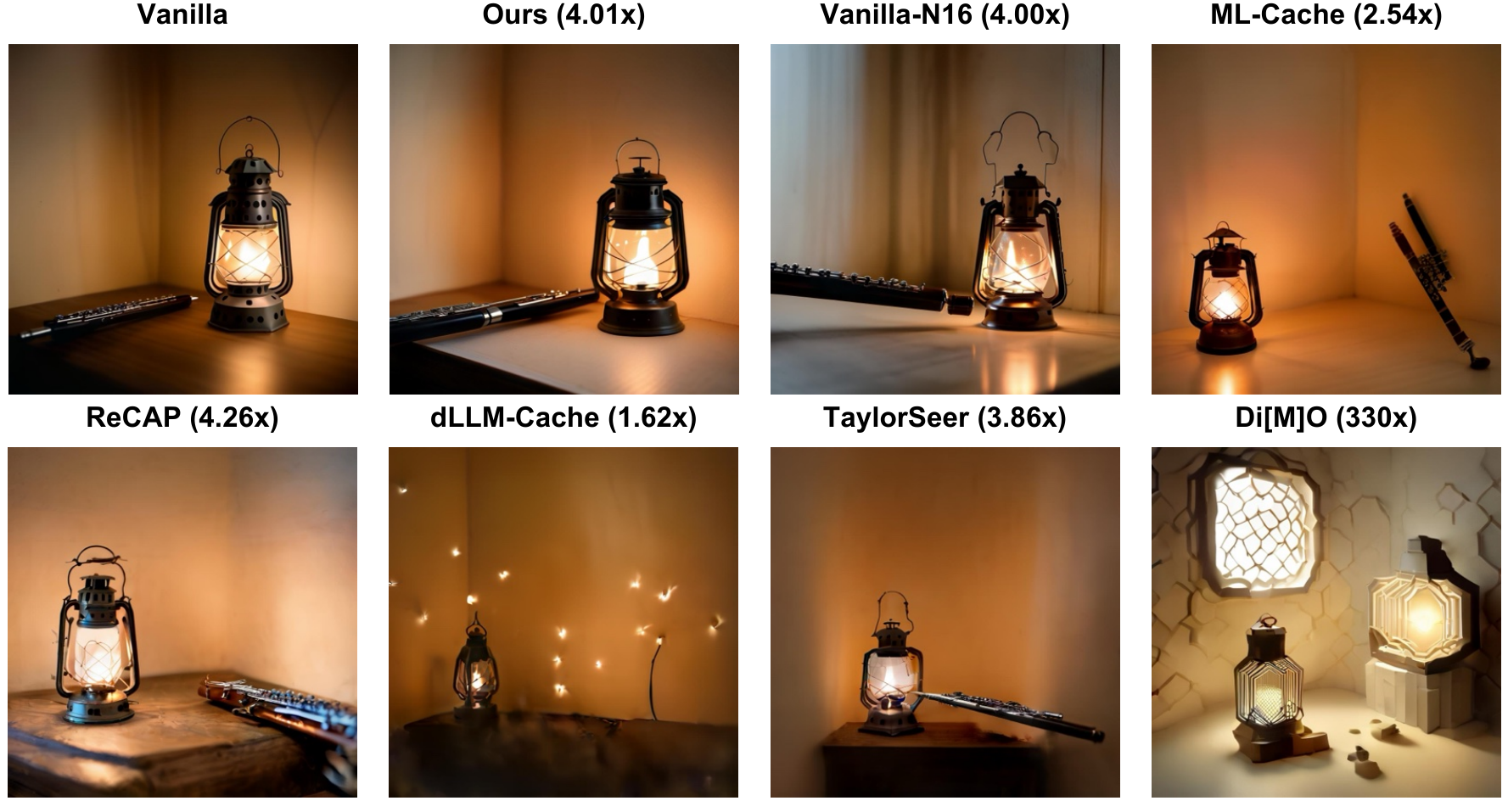}
    \caption{Prompt: In an otherwise silent room, the melodic rhythm of a recorder being played gently reverberates throughout the space. A single lantern with a hexagonal frame and intricate metalwork sits nearby, emitting a soft, warm light that flickers subtly, illuminating the immediate vicinity. The walls, which are painted a muted cream color, catch the lantern's glow, creating a cozy ambiance around the musician.}
    \label{supp:fig:pic7}
\end{figure*}

\begin{figure*}
    \centering
    \includegraphics[width=0.91\linewidth]{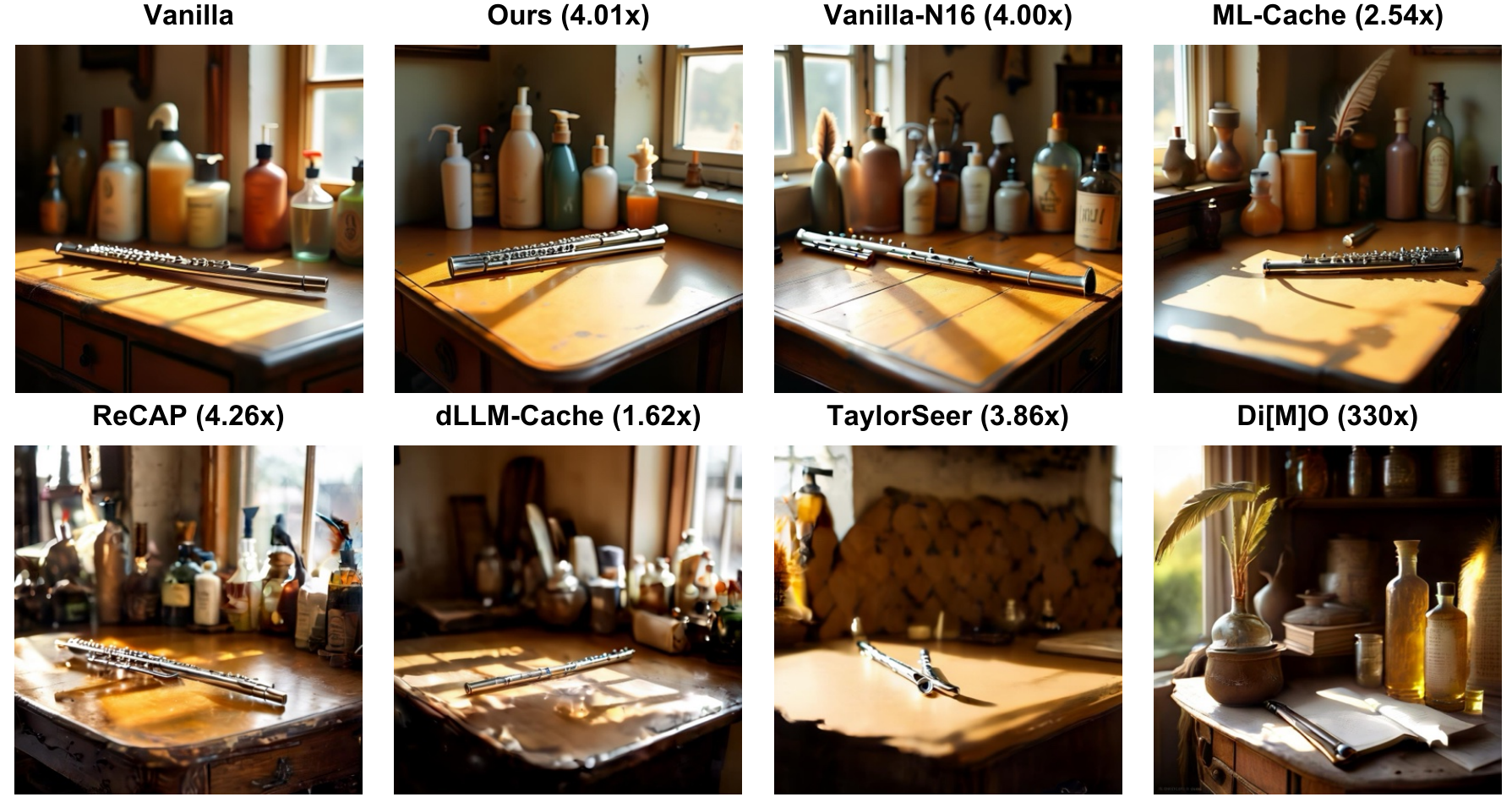}
    \caption{Prompt: A rustic, vintage wooden desk sitting in the corner of an antique store, its surface aged to a warm honey color. Upon the desk lies a silver flute, its polished surface gently reflecting the soft, golden light from the late afternoon sun streaming through a nearby window. Behind the flute, an array of eclectic cleaning products, from old-fashioned feather dusters to vintage glass bottles, is artfully arranged, adding to the charm of the setting. These items catch the sunlight in a way that casts an array of subtle shadows and highlights across the desk's surface.}
    \label{supp:fig:pi8}
\end{figure*}

\begin{figure*}
    \centering
    \includegraphics[width=0.91\linewidth]{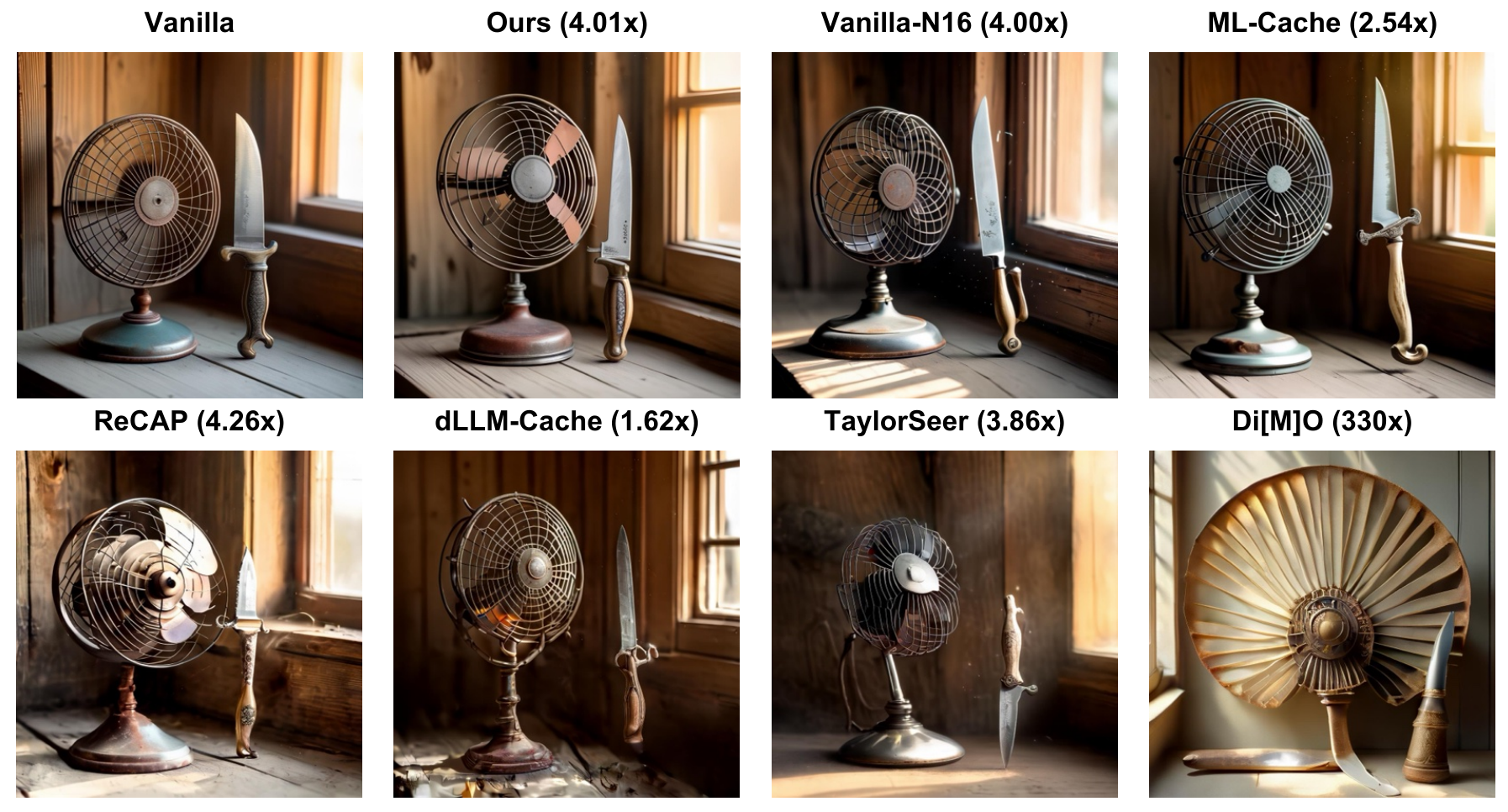}
    \caption{Prompt: An old fashioned, metallic fan with a rounded base and rusted blades stands next to an ornate, antique knife with a weathered bone handle. Both items are propped against a timeworn wooden wall that bears the marks of age and the warm, golden hue of the afternoon sunlight filtering through a nearby window. The dust particles in the air catch the light, highlighting the vintage aura of the scene.}
    \label{supp:fig:pic9}
\end{figure*}